\documentclass[10pt,twocolumn]{article}
\usepackage[margin=0.8in,columnsep=0.25in]{geometry}
\usepackage{amsmath,amssymb}
\usepackage{booktabs}
\usepackage{graphicx}
\usepackage{multirow}
\usepackage{xcolor}
\usepackage{microtype}
\usepackage[hidelinks]{hyperref}
\usepackage{url}
\usepackage{caption}
\usepackage{placeins}

\usepackage{needspace}

\newcommand{\ours}{Fathom}

\newcommand{\fig}[3]{\IfFileExists{figs/#1}{\includegraphics[width=#2]{figs/#1}}{\fbox{\parbox{#2}{\centering\small missing figure \detokenize{#1}: #3}}}}
\newcommand{\tabin}[1]{\IfFileExists{tables/#1.tex}{\input{tables/#1}}{\fbox{\small missing table \detokenize{#1}}}}
\graphicspath{{figs/}}
\newcommand{\activeFortyEightMax}{34}
\newcommand{\activeFortyEightMin}{30}

\newcommand{\agentCalibOracleSame}{40}

\newcommand{\agentCtxMean}{95}

\newcommand{\agentN}{40}
\newcommand{\agentNBig}{20}

\newcommand{\agentOsimBiglandmark}{0.47}
\newcommand{\agentOsimBigplanesK}{0.67}

\newcommand{\agentOsimBigsparq}{0.49}

\newcommand{\agentOsimplanesK}{0.53}
\newcommand{\agentOsimplanesKeighty}{0.60}

\newcommand{\agentOsimplanesSeighty}{0.64}

\newcommand{\agentOsimsparq}{0.49}
\newcommand{\agentOsimsparqthirtytwo}{0.60}

\newcommand{\agentPairedAgentEightyWiki}{+0.07}
\newcommand{\agentPairedAgentFortyEightWiki}{+0.01}

\newcommand{\agentPairedBigSparq}{+0.18}
\newcommand{\agentPairedDsAgentWiki}{-0.01}
\newcommand{\agentPairedEightySparqThirtyTwo}{-0.00}

\newcommand{\agentPairedSEBigSparq}{0.05}

\newcommand{\agentPairedSEEightySparqThirtyTwo}{0.04}

\newcommand{\agentPairedSESelfEightySparqThirtyTwo}{0.04}

\newcommand{\agentPairedSESparq}{0.05}
\newcommand{\agentPairedSelfEightySparqThirtyTwo}{+0.04}

\newcommand{\agentPairedSparq}{+0.05}

\newcommand{\agentWinsBigSparq}{14}

\newcommand{\agentWinsEightySparqThirtyTwo}{20}

\newcommand{\bTwoRatioLong}{1.63}
\newcommand{\bTwoRatioShort}{1.44}

\newcommand{\calibAgentEightyFiveTwelve}{0.0014}

\newcommand{\calibAgentFortyEightFiveTwelve}{0.0053}

\newcommand{\calibDsAgentFiveTwelve}{0.0044}

\newcommand{\calibDsWikiFiveTwelve}{0.0050}

\newcommand{\calibWikiEightyFiveTwelve}{0.0017}

\newcommand{\calibWikiFortyEightFiveTwelve}{0.0057}

\newcommand{\fortyBytesFewerPct}{31}
\newcommand{\fortyEightBytesFewerPct}{18}
\newcommand{\hbmLandmark}{44}
\newcommand{\hbmScanMax}{48}
\newcommand{\hbmScanMin}{45}
\newcommand{\hbmThumb}{55}

\newcommand{\hthFortyN}{7}

\newcommand{\hthFortyWins}{6}
\newcommand{\hthRatioLong}{5.3}
\newcommand{\hthRatioMax}{5.3}
\newcommand{\hthRatioMin}{1.1}
\newcommand{\issuechanrsixteen}{24}
\newcommand{\issuechanr}{14}
\newcommand{\issuelandmark}{22}

\newcommand{\issueplanesmean}{48}
\newcommand{\issueplanesmeansixtyfour}{15}
\newcommand{\issueplanesthumb}{29}
\newcommand{\kernelBytesFewerPct}{38}
\newcommand{\kernelRatioLong}{0.70}
\newcommand{\kernelRatioShort}{0.73}
\newcommand{\kratioKltFortyEight}{0.0196}
\newcommand{\kratioRawFortyEight}{0.0300}
\newcommand{\marginHighRatio}{4.4}
\newcommand{\marginLowRatio}{1.6}
\newcommand{\matchDSmax}{74}
\newcommand{\matchDSmin}{46}
\newcommand{\matchDSratioMax}{2.9}
\newcommand{\matchDSratioMin}{1.8}
\newcommand{\matchLokiMax}{75}
\newcommand{\matchLokiMin}{56}
\newcommand{\matchLokiN}{4}
\newcommand{\matchSQmax}{92}
\newcommand{\matchSQmin}{38}

\newcommand{\nSettings}{7}

\newcommand{\pcieGatherMax}{26.2}
\newcommand{\pcieGatherMin}{25.1}
\newcommand{\pcieGatherWorst}{23.7}
\newcommand{\pcieMemcpyMax}{25.9}
\newcommand{\pcieMemcpyMin}{21.6}
\newcommand{\pciePercallMax}{22.4}
\newcommand{\pciePercallMin}{0.4}
\newcommand{\ratioChanOverForty}{1.77}
\newcommand{\ratioGPUchanrsixteen}{1.05}
\newcommand{\ratioGPUchanr}{1.67}
\newcommand{\ratioGPUchanAt}{1.37}
\newcommand{\ratioGPUlandmark}{2.50}

\newcommand{\ratioGPUplanesmeansixtyfour}{1.21}
\newcommand{\ratioGPUplanesthumb}{3.12}
\newcommand{\ratioSparqSixteenOverForty}{1.11}

\newcommand{\ratioWallchanr}{1.38}
\newcommand{\ratioWalllandmark}{2.07}

\newcommand{\ratioWallplanesthumb}{2.59}
\newcommand{\realRatio}{1.26}
\newcommand{\realRatioSparq}{1.00}
\newcommand{\riseDS}{6}

\newcommand{\riseOurs}{15}
\newcommand{\riseSparq}{5}
\newcommand{\rowsMBmax}{216}
\newcommand{\rowsMBmin}{167}

\newcommand{\rulerLandmarkGapLong}{0.130}
\newcommand{\rulerLandmarkGapShort}{0.043}
\newcommand{\rulerLongN}{20}
\newcommand{\rulerN}{40}

\newcommand{\rulerSELong}{0.034}
\newcommand{\rulerSEShort}{0.014}
\newcommand{\rulerScanDevLong}{0.025}
\newcommand{\rulerScanDevShort}{0.008}

\newcommand{\scanShareLong}{24}
\newcommand{\scanShareShort}{16}
\newcommand{\scanShareOursLong}{30}
\newcommand{\scanShareOursShort}{21}
\newcommand{\sharedPctChan}{20}
\newcommand{\sharedPctOurs}{33}

\newcommand{\sparqSixteenBytesMorePct}{22}
\newcommand{\sparqSixteenFlatWins}{6}
\newcommand{\thumbMatchMax}{148}
\newcommand{\thumbMatchMin}{28}

\newcommand{\wallGapSparqPct}{7}

\title{Fathom: Per-Query Read Depth for Sparse Decoding over Offloaded KV Caches}
\author{Vivek Kalyanarangan}
\date{}

\begin{document}
\twocolumn[
\maketitle
\begin{center}\begin{minipage}{0.92\textwidth}
\begin{abstract}
When agentic sessions run to a million tokens with many sessions resident at once, the KV cache and the index that ranks it live in host memory, and the scan that ranks all $n$ keys for a top-$k$ step becomes the traffic that bounds decoding. We present \ours, a key scan in which each query decides how many bits of each key channel to read. The 4-bit K cache is stored channel-major as bit planes, so a prefix of $t$ planes is exactly the channel's $t$-bit quantizer, and the query spends its bit budget by reverse water-filling over the variance-weighted importance of its channels. At one million tokens on Qwen3-8B a decode step is \ratioGPUchanr$\times$ faster in GPU time than with the 136-bit scans of Double Sparsity, Loki and SparQ $r=32$, and in the same GPU time as SparQ's 68-bit read ($r=16$) \ours{} reads \fortyEightBytesFewerPct\% fewer bytes with lower attention error on six of seven model and context settings. On RULER-style tasks every per-token scan matches exact top-$k$ decoding, and on real coding-agent sessions \ours{} reaches the step agreement of the most accurate 136-bit scan at 92 bits. The store is the 4-bit K copy a quantized serving stack already holds, and the method is not faster when the index is resident in GPU memory.
\end{abstract}

\begin{center}\small Code and results: \url{https://github.com/vivekkalyanarangan30/fathom}\end{center}

\end{minipage}\end{center}
\vspace{1.2em}
]

\section{Introduction}
A coding or browsing agent carries a context of hundreds of thousands of tokens for hours, most of it cached rather than newly generated, and a server hosts many such sessions at once. Their KV caches no longer fit beside the weights in GPU memory and are held in host memory or a slower tier, so a decode step is bounded by what crosses the interconnect. Decoding was already bandwidth-bound, since each generated token reads the model weights once and, at long context, the entire KV cache once per layer. Offloading makes the cache read the dominant term. Grouped-query attention (GQA)~\cite{gqa} and 2- to 4-bit KV quantization~\cite{kivi,kvquant} shrink the cache. Top-$k$ sparse attention~\cite{loki,doublesparsity,sparq,quest} shrinks the \emph{read}, fetching only the $k$ keys and values with the highest attention scores. Since the scores are unknown before the read, every such method first scans a cheap representation of all $n$ keys to rank them. The scan is a stream of $n$ small records whose cost is measured in bits per token, and it grows with $n$ while the fetch of the winners does not.

The numbers for Qwen3-8B (36 layers, 8 KV heads of 128 channels, 4 query heads per KV head) at 32k tokens illustrate the split. Dense attention reads the whole bf16 KV cache, 4.8\,GB per step. Top-$k$ with $k=512$ per query head fetches the union of the four heads' winners, at most 2048 rows per KV head and about 200\,MB per step in our runs, a cost that does not grow with the context. A Loki, Double Sparsity or SparQ ($r=32$) scan reads 32 coordinates of each key at 4 bits plus a block scale, 136 bits per token for each of the 288 layer--head pairs, 557\,KB per pair and 160\,MB per step at 32k. That term grows linearly with context, to 5.1\,GB at one million tokens, while the winner rows stay near 200\,MB. All bit counts in this paper are per token per KV head per layer, under the one convention of \S\ref{sec:setup}.

Two levers reduce the scan. One scores fewer things, such as pages, blocks or landmarks~\cite{quest,infllm,shadowkv}. The other reads fewer bits per key. Existing per-token scans fix the bits in advance. Loki~\cite{loki} reads $r$ principal-component coordinates of every key; Double Sparsity~\cite{doublesparsity} reads $c$ channels chosen offline from a 4-bit label cache; SparQ~\cite{sparq} lets the query pick the $r$ channels with the largest summed $|q|$ over the query heads of the group and reads them at full depth; thumbnail scans~\cite{ffd} read every channel at 2 bits. In all of them the depth of the read is the same for every channel the method touches.

This paper lets the read depth be decided per query and per channel. Two observations make this possible. First, quantization error falls by $4\times$ per bit. If channel $j$ of a key has been read to a depth of $t_j$ bits, one more bit reduces the remaining score error by an amount proportional to $g_j4^{-t_j}$, where $g_j$ is that channel's contribution to the score variance for this query. The first bit of an important channel is worth far more than its fourth, and the fourth bit of an important channel can be worth less than the first bit of a minor one. Allocating bits in order of this marginal value is reverse water-filling~\cite{gershogray}, and it has a closed form. Second, if the 4-bit K cache is stored as bit planes, channel-major, the first $t$ planes of a channel \emph{are} its $t$-bit mid-rise quantizer with the same block scale, so a prefix read is exact and contiguous and no lower-precision copy is needed. Together they turn a 4-bit copy of the keys into a multi-resolution index that each query reads to the depth it needs.

We evaluate \ours{} against Loki, Double Sparsity, SparQ, a 2-bit thumbnail and a block-landmark index on \nSettings{} model and context settings, on RULER-style tasks at 32k and 128k, and in the hardware regime the method is built for as well as the ones it is not. The byte saving holds on every setting; it becomes a time saving when the scan bytes cross PCIe, and \S\ref{sec:analysis} analyses when that is.

\paragraph{Contributions.}
\begin{itemize}\setlength\itemsep{1pt}
\item A measured result in the regime that motivates the work: with the KV cache and its index in host memory, a decode step is \ratioGPUchanAt$\times$ faster at 256k tokens and \ratioGPUchanr$\times$ at 1M in GPU time than with the 136-bit scans, at equal or better accuracy than Double Sparsity and Loki; against SparQ's 68-bit read it reads \fortyEightBytesFewerPct\% fewer bytes in the same GPU time and is \hthRatioMin--\hthRatioMax$\times$ more accurate (\S\ref{sec:offload}, \S\ref{sec:h2h}).
\item A bit-plane key store in which prefix reads are exact lower-precision quantizers, and a per-query reverse water-filling rule for the read depth of every channel, with an optional per-layer budget calibrated once (\S\ref{sec:method}).
\item Equal-error byte savings of \matchDSratioMin--\matchDSratioMax$\times$ against Double Sparsity's 136-bit scan on \nSettings{} settings up to 128k tokens, downstream parity with the exact top-$k$ oracle on RULER-style tasks, and on real coding-agent sessions the step agreement of the most accurate 136-bit scan at 92 bits (\S\ref{sec:fidelity}, \S\ref{sec:ruler}, \S\ref{sec:agent}).
\item A basis rule that uses raw channels for models with QK-norm (query and key normalisation, as in Qwen3) and planes of the Karhunen--Lo\`eve transform (KLT) of the keys otherwise (Llama-3.1, Qwen2.5) (\S\ref{sec:basis}).
\item An analysis of the HBM-resident case: with the index in GPU high-bandwidth memory (HBM) the method is not faster, and an arithmetic-per-byte analysis says why (\S\ref{sec:analysis}), with ablations of every design choice (\S\ref{sec:abl}, Appendix~\ref{app:abl}).
\end{itemize}

\paragraph{Summary of measured results.} \ours{} is built for one situation: sparse decoding when the KV cache and the scan index are in host memory. Table~\ref{tab:regimes} summarises the measurements in that regime and in the regimes where the method does not help.
\begin{table*}[t]\centering\small
\caption{Measured results by regime on an A100 (Qwen3-8B, $k=512$). GPU time per decode step; the 136-bit scans are Double Sparsity, Loki and SparQ $r=32$.}
\label{tab:regimes}
\begin{tabular}{p{5.2cm}p{3.4cm}p{8.2cm}}
\toprule
regime & what binds & \ours{} (56-bit read unless noted) \\
\midrule
KV rows and index in host memory, 1M tokens & PCIe bytes & \textbf{\ratioGPUchanr$\times$} faster than the 136-bit scans, \textbf{\ratioGPUlandmark$\times$} than landmarks; the same GPU time as SparQ $r=16$ at 56 bits (ratio \ratioGPUchanrsixteen) and \textbf{\ratioSparqSixteenOverForty$\times$} faster at 47 bits, at lower error \\
same, 128k tokens, real prefill & PCIe bytes & \textbf{\realRatio$\times$} faster than the 136-bit scans; SparQ $r=16$ takes \realRatioSparq$\times$ its GPU time \\
real coding-agent sessions, 100k tokens, $k=2048$ & scan fidelity & step agreement with exact top-$k$ \textbf{\agentOsimBigplanesK}, SparQ $r=16$ \agentOsimBigsparq, landmarks \agentOsimBiglandmark \\
real coding-agent sessions, 100k tokens, $k=512$ & scan fidelity & \textbf{\agentOsimplanesKeighty} at 92 bits, equal to SparQ $r=32$'s \agentOsimsparqthirtytwo{} at 136 bits \\
rows in host memory, index in HBM & the shared row fetch & not faster; all per-token scans land at \hbmScanMin--\hbmScanMax\,ms \\
everything in HBM & arithmetic per scanned bit & not faster; the scan kernel is $1.4\times$ slower than a nibble scan \\
\bottomrule
\end{tabular}
\end{table*}

\section{Background and Related Work}
\paragraph{Top-$k$ decoding.} Given a query $q\in\mathbb{R}^D$ and keys $K\in\mathbb{R}^{n\times D}$, exact attention weights are $\mathrm{softmax}(qK^\top/\sqrt D)$. Attention mass concentrates on few keys, so H$_2$O~\cite{h2o} and StreamingLLM~\cite{streamingllm} evict, while Loki, Double Sparsity, SparQ and Quest keep the full cache and select at decode time by approximate score. We evaluate every method under one protocol: the first 4 tokens and the last 32 are always kept, following StreamingLLM and H$_2$O, and the remaining $k-36$ are the top-scoring keys under the method's scores, per query head; each method is measured by the exact attention output over its selected set. SparQ's reallocation of the unselected attention mass onto a mean value is not applied to any method.

\paragraph{Loki.} Loki~\cite{loki} rotates keys into the principal-component (PCA) basis of calibration keys and scores queries against the first $r$ coordinates, kept in the model's precision; the original method stores no second copy of the keys and its $r=32$ read is 512 bits per token. To compare byte for byte with 4-bit scans we quantize the $r$ coordinates to 4 bits with the same block scales as every other method (136 bits at $r=32$), and we order the basis by calibration score variance rather than by eigenvalue, which favours Loki. Its accuracy is rank-limited: on Qwen3-8B the 32-coordinate basis misses late-layer directions at 32k and fp16 coordinates do not repair it (\S\ref{sec:abl}).

\paragraph{Double Sparsity.} Double Sparsity~\cite{doublesparsity} keeps a label cache of $c$ channels per head chosen offline from a calibration statistic of the score contribution and scans it at 4 bits. We choose channels by the per-head product of mean $|q|$ and mean $|k|$ on calibration text and run $c=32$, a quarter of the channels; the paper's own default is a sixteenth, eight channels, and it also reports a half and all of them. At $c=32$ the label cache reads 136 bits per token with fp16 block scales, the same as the other fixed-depth scans, and is 17 bytes per token stored beside the K cache. It is the strongest offline baseline on the Qwen3 models; Loki is stronger on the other three models.

\paragraph{SparQ.} SparQ~\cite{sparq} lets each query choose the $r$ channels with the largest $|q_j|$ and reads them from a channel-major K store. Under grouped-query attention its published rule sums $|q|$ over the query heads that share a KV head before the top-$r$, so one set of $r$ channels is read per KV head. SparQ's own store is fp16 (512 bits at $r=32$); with the 4-bit codes we give every scan (\S\ref{sec:setup}) that is 68 bits at $r=16$ and 136 at $r=32$. We implement that rule. A stronger variant that lets every query head pick its own channels and reads the union of the picks is not SparQ's method; we report it once as an ablation (\S\ref{sec:abl}). SparQ also sums the approximate scores over the group before its top-$k$; we do not adopt that for any method and select top-$k$ per query head throughout.

\paragraph{Block and landmark selection.} Quest~\cite{quest} scores 16-token pages by per-channel minimum and maximum keys; ShadowKV~\cite{shadowkv} scores 8-token chunks by their mean key, keeps a low-rank K cache on the GPU and offloads V; InfLLM~\cite{infllm} scores blocks by representative tokens. This is the ``score fewer things'' lever, orthogonal to ours. Our landmark baseline follows ShadowKV, the mean key of each 8-token block in fp16, 256 bits per token, run at the paper's $k$ rather than at the block budgets those systems use. Both systems keep their landmark index in GPU memory; the host-memory setting of \S\ref{sec:offload} is ours, not theirs.

\paragraph{Thumbnail scans.} FFD~\cite{ffd} splits K into a 2-bit thumbnail and an 8-bit residual, selects by a score threshold and fuses scan and attention in one kernel. Our 2-bit thumbnail baseline reads all 128 channels at two planes (288 bits per token) and selects top-$k$ with exact attention on the selected set, which is more generous than FFD's own scoring; the bit-plane store makes it a special case of ours with uniform depth.

\paragraph{The store as the index.} Self-Indexing KVCache~\cite{selfindex} makes the same structural argument, that a compressed copy of the keys can serve as the selection index with no separate structure, and realises it with a sign-based 1-bit vector quantizer, one fixed depth for every channel and every query. Louver~\cite{louver} builds an index that answers a threshold query over the keys with no false negatives, again at a fixed representation. Our difference is the depth itself: a prefix of the bit planes is an exact quantizer, so depth becomes a per-query, per-channel decision rather than a design-time constant, and A4 in \S\ref{sec:abl} measures what that freedom is worth against uniform depth at equal bytes.

\paragraph{KV quantization.} KIVI~\cite{kivi} (2-bit, per-channel keys and per-token values), KVQuant~\cite{kvquant} (non-uniform, pre-RoPE, dense-and-sparse) and TurboQuant~\cite{turboquant} (random rotation, Lloyd--Max scalar quantizer and a 1-bit residual) compress the cache itself; Any-Precision LLM~\cite{anyprecision} stores weights in bit planes so that a prefix is a lower-precision model. Our layout applies the bit-plane idea to the K cache and reads a per-query prefix per channel.

\paragraph{Offloaded retrieval.} MagicPIG~\cite{magicpig} samples with locality-sensitive hashing (LSH) tables on the CPU, RetroInfer~\cite{retroinfer} and RetrievalAttention~\cite{retrievalattention} keep vector indexes in host memory, and InfiniGen~\cite{infinigen} prefetches speculatively from a host-resident cache. This is the regime where scan bytes are time and where our gains are measured.

\section{Method}\label{sec:method}
\subsection{Cost model}
Let $N$ be the number of sequences decoding together, each with a context of $n$ tokens (so $Nn$ tokens are resident), $b$ the scan bits per token per KV head, $H$ the KV heads, $L$ the layers, $G$ the query heads per KV head, $k_\cup\le Gk$ the number of distinct winner rows per KV head after the union over the group's picks, and $R$ the bytes of a full K$+$V row. A decode step reads
\begin{equation}
\text{bytes} \approx W + N\,L\,H\,\Big(\tfrac{b}{8}\,n + k_\cup R\Big),
\end{equation}
where $W$ is the weight read once per step. The scan term grows with the resident tokens $Nn$; the row term grows with $N$ only, since each sequence fetches at most $k_\cup$ rows per head whatever its length. Reducing $b$ matters exactly when $b\,n/8$ is the largest of the three terms, which is the long-context, many-session regime this paper targets.

\subsection{Bit-plane K cache}
Keys are quantized once to 4 bits with a uniform, symmetric code per channel and one fp16 scale per 64-token block. Uniformity is what makes a prefix of bit planes an exact coarser quantizer; a non-uniform code would not have this property. For block $\beta$ and channel $j$ let $a_{\beta,j}=\max_{i\in\beta}|k_{i,j}|$ be the block maximum and $s_{\beta,j}=a_{\beta,j}/8$ the cell width; the code is
\begin{equation}
c_{i,j}=\mathrm{clip}\big(\lfloor k_{i,j}/s_{\beta,j}\rfloor,-8,7\big)+8\in[0,16),
\end{equation}
so that code 8 is zero and the most significant bit is the sign.

\begin{figure*}[t]\centering
\fig{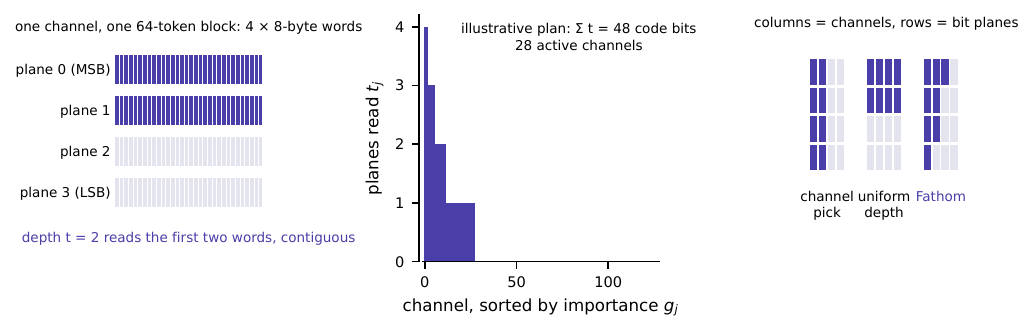}{\textwidth}{layout schematic}
\caption{Left: one channel of one 64-token block as four bit planes; a depth-$t$ read is the first $t$ words. Middle: an illustrative per-query read of a 128-channel key at a mean of 48 bits, a staircase over channels sorted by importance $g_j$ (the measured number of active channels is \activeFortyEightMin--\activeFortyEightMax, Table~\ref{tab:active}). Right: the read shapes of channel picks (Double Sparsity, SparQ), uniform depth (thumbnails) and ours.}
\label{fig:method}
\end{figure*}

Bit $p$ ($0$ = most significant) of the 64 codes of block $\beta$, channel $j$, is packed into one 64-bit word $P_{\beta,j,p}$, the plane. Planes are stored channel-major, so the words $P_{\cdot,j,0\ldots t-1}$ are contiguous over the sequence. Reading the first $t$ planes of a channel gives $c^{(t)}=c\gg(4-t)$, and the dequantized value $(c^{(t)}+\tfrac12-2^{t-1})\,a_{\beta,j}/2^{t-1}$ is exactly the $t$-bit mid-rise quantizer of the range $[-a_{\beta,j},a_{\beta,j})$ with cells of width $a_{\beta,j}/2^{t-1}$; at $t=4$ it reduces to $(c-8+\tfrac12)\,s_{\beta,j}$. The store is a 4-bit copy of K with fp16 block scales, 68 bytes per token per KV head. In a serving stack that keeps its K cache in 4-bit form the planes can serve as that cache; our experiments keep bf16 rows for the winners and treat the planes as a separate index, and the memory comparison in \S\ref{sec:analysis} charges the full 68 bytes. Figure~\ref{fig:method} shows the layout.

\paragraph{Compatibility with other KV quantizers.} The layout requires a scalar code per channel whose bit prefixes are coarser quantizers; it is not agnostic to the quantization family. A fixed rotation before quantization is compatible, and \S\ref{sec:basis} uses one, but the rotation must concentrate variance (a KLT) rather than spread it (the random rotation of TurboQuant~\cite{turboquant}), because water-filling has nothing to allocate when every coordinate carries equal variance; TurboQuant's Lloyd--Max quantizer is in addition non-uniform. Uniform integer KV formats, including KIVI's per-channel codes with a zero point~\cite{kivi} and INT8, are compatible directly, since a prefix of a uniform code is a coarser uniform code. Non-uniform codes lose the exactness property. KVQuant's lookup-table datatypes~\cite{kvquant} would need their table stored in value order and a per-depth table, and its keys quantized before the rotary position embedding (RoPE) and its separate outlier component would each need handling we have not built; FP8, whose prefixes are sign and exponent bits, is monotone in magnitude and would likewise need measured rather than $4^{-t}$ marginal gains. Codebook vector quantization has no per-channel bit depth; residual VQ, being progressive by stage, would admit a per-query depth in stages, which we do not explore.

\subsection{Per-query read depth}
For the group $\{q^{(h)}\}_{h}$ of query heads sharing one KV head, the contribution of channel $j$ to the scores has variance across keys proportional to
\begin{equation}
g_j=\sum_{h}\big(q^{(h)}_j\big)^2\,\mathrm{Var}(k_j),
\label{eq:g}
\end{equation}
with $\mathrm{Var}(k_j)$ measured once on calibration keys. Reading $t$ planes of a channel is a $t$-bit uniform quantizer over $[-a_{\beta,j},a_{\beta,j}]$: it splits that range into $2^t$ cells of width $\Delta=2a_{\beta,j}/2^t$, and a value is off from its cell centre by an error of variance $\Delta^2/12=a_{\beta,j}^2/(3\cdot4^t)$. Every extra plane therefore divides the error variance of that channel by four. That error enters the score multiplied by $q_j$, so summed over the heads of the group, and with $a_{\beta,j}^2$ standing in for $\mathrm{Var}(k_j)$, channel $j$ contributes an expected squared score error proportional to $g_j4^{-t_j}$, and the total is $\sum_j g_j4^{-t_j}$. Minimizing it under $\sum_j t_j\le B$ is reverse water-filling~\cite{gershogray}: bits go first to the channels with the largest $g_j$, and each channel's depth is set by how far its importance sits above a common water line $\theta$. With integer depths the optimum is
\begin{equation}
t_j=\mathrm{clip}\big(\mathrm{round}(\log_4(g_j/\theta)),\,0,\,4\big),\qquad \sum_j t_j\le B,
\label{eq:wf}
\end{equation}
Each $t_j$ is a step function of $\theta$ that falls as $\theta$ rises, so $\sum_j t_j$ is too; the water line is the smallest $\theta$ at which the sum fits the budget, and 30 bisection steps on $\log\theta$ per query group locate it. Channels with $t_j=0$ are skipped entirely and the rest are the active channels; a read at mean 48 bits touches \activeFortyEightMin--\activeFortyEightMax{} of 128 channels at 1--4 planes each on the models we test (Table~\ref{tab:active}). The plan is shared by the $G$ heads of the group, so the K bytes are read once for all of them. Two choices differ from SparQ: $g_j$ weights the query by the key variance rather than ranking by $|q_j|$, and depth is graded rather than all-or-nothing. Appendix~\ref{app:toy} works Eq.~\ref{eq:wf} through a six-key example.

\subsection{Per-layer budgets}
Layers differ in how peaked their score distributions are. The flat budget gives every layer the same $B$. A greedy allocation on calibration text instead assigns budgets $B_\ell\in\{24,\dots,128\}$ to layers at a target mean (48 or 64 bits), each step moving bits to the layer with the largest error drop per bit; we call this the per-layer plan. On the Qwen3 models it lowers error at mean 48 by about half at 16k and by 8--11\% at 32k, and is worse at mean 64 on Qwen3-8B at 32k with $k=128$; it is worse on Qwen2.5-7B and Llama-3.1-8B (up to $1.7\times$ and $2.1\times$ at mean 64) and neutral to better on Qwen2.5-7B-1M, and a plan calibrated at 16k evaluated at 32k is worse than the flat budget (Tables~\ref{tab:qwen3_8b} and \ref{tab:plan}, \S\ref{sec:abl}). Our recommendation is a flat budget by default: it needs no calibration and has no context-length dependence. The per-layer plan is a tuning step for a fixed deployment, calibrated at that deployment's context length; we report both throughout.

\subsection{Basis rule}\label{sec:basis}
Water-filling over raw channels assumes score variance is concentrated in few channels. On models with QK-norm (Qwen3) it is, and rotating keys with the calibration KLT spreads the per-query sparsity and raises error. On models without QK-norm (Llama-3.1-8B, Qwen2.5-7B, Qwen2.5-7B-1M) the same rotation lowers error at equal bits (Table~\ref{tab:basis}). The rotated variant stores planes of $(k-\mu)V$ with the score-variance-ordered eigenbasis $V$ of the calibration keys, per KV head and layer, and uses $g_j=\sum_h((q^{(h)}V)_j)^2\lambda_j$; nothing else changes. Unlike Loki nothing is truncated; all $D$ rotated coordinates are stored, and the query decides how deep to read each. The rule is decided once per model by evaluating both stores on calibration text, and every result below applies it: raw planes on Qwen3, rotated planes on the other three models.

\subsection{Reading a host-resident store}\label{sec:gather}
When the store lives in host memory, a GPU kernel reading mapped host memory word by word is limited by the small, scattered transactions, whereas a contiguous copy runs at the link rate (Fig.~\ref{fig:pcie}). With the whole sequence as one channel-major block, the first $t_j$ planes of channel $j$ are one contiguous run of $t_j\,n/8$ bytes. A gather kernel copies one run per active channel, plus that channel's scales, into a staging buffer, and the scan runs in HBM; the run list has a fixed size, one slot per channel with inactive slots of length zero, so no host synchronization is needed. The same transfer is given to every baseline in the offload experiments.

\section{Experimental Setup}\label{sec:setup}
\paragraph{Models and hardware.} Qwen3-8B (16k and 32k), Qwen3-4B (16k)~\cite{qwen3}, Qwen2.5-7B (32k) and Qwen2.5-7B-Instruct-1M (32k and 128k), with activations captured, fidelity computed, RULER-style tasks run and all timing measured on one NVIDIA A100-SXM4-80GB pod (PCIe 4.0, 2\,TB host RAM). Llama-3.1-8B~\cite{llama3} at 4k uses activations captured on an NVIDIA L4 and evaluated with the same fidelity code. Calibration uses the Wikitext-103 train split and evaluation the test split, both at the evaluated context length.

\paragraph{Fidelity metric.} For 128 decode positions at the end of the window (64 at 128k), all layers and KV heads, the selected set is sink 4 + local 32 + top-$(k-36)$ by the method's scores, per query head; we report the mean relative $L_2$ error of the attention output from that set against dense attention. $k$ scales with context (128 at 4k, 256 at 16k, 512 at 32k) unless stated; \S\ref{sec:ratio} varies it. Each setting is one held-out Wikitext window.

\paragraph{RULER-style tasks.} These synthetic retrieval and state-tracking tasks are our proxy for the long-range recall that long agent sessions depend on; they share structure with coding and tool-use transcripts, not content, and \S\ref{sec:limits} says what a workload-level evaluation would need. Tasks with RULER templates~\cite{ruler} on a Wikitext haystack: single-needle, multi-key, multi-value and multi-query needle-in-a-haystack, variable tracking and frequent-word extraction at 32k (Qwen3-8B, $k=128$, \rulerN{} samples per task), and multi-key, multi-query, variable tracking and frequent-word extraction at 128k (Qwen2.5-7B-Instruct-1M, $k=128$, \rulerLongN{} samples). One dense prefill per sample, then greedy decoding branched per method from the same KV cache with sparse attention on every generated token. Score is the fraction of gold strings present in the generation. The decode scorers reproduce the fidelity code to $10^{-4}$ and the dense path reproduces HuggingFace generation token for token, checked on this pod before the runs.

\paragraph{Offload harness.} Model weights on the GPU; K/V rows and, unless noted, the scan store in pinned host memory. Each step runs the scan, the top-$k$ per query head, the gather of the union of the selected rows over PCIe, and exact attention. The primary timing metric is GPU time, the profiler's kernel plus memcpy time of a decode step, because it measures the work the method changes. Wall-clock is the median of the steps after two warm-up steps. It adds the host-side time of this research harness, which differs by method and would not exist in a fused implementation. Table~\ref{tab:offload} reports both and Table~\ref{tab:hbm} GPU time. The synthetic runs of Table~\ref{tab:offload} and Figure~\ref{fig:offload} use a synthetic KV cache (calibration keys tiled to $n$, random values), because no model here prefills a million tokens, and are timing only; the 32k--128k runs of Table~\ref{tab:hbm} use a real prefill.

\paragraph{Baselines.} Double Sparsity $c=32$, Loki $r=32$ and $r=64$, SparQ $r=16$ and $r=32$ under its grouped-query rule, a 2-bit all-channel thumbnail, the ShadowKV-style block-mean landmark index with 8-token blocks, the exact top-$k$ oracle and dense attention. All scans use 4-bit codes with identical block scales. In the timing harness the 16- and 32-channel scans select their channels per query by the group's $\sum|q|$, so the 32-channel row stands for SparQ $r=32$ and for the Loki and Double Sparsity byte count at once.

\paragraph{Byte accounting.} A single convention applies throughout. A method's bits per token are the code bits it reads plus 16 bits per 64 tokens for the fp16 block scale of every channel it reads. That gives 136 for 32 channels at 4 bits, 68 for 16, 288 for the 2-bit thumbnail, 544 for the full 4-bit scan and 256 for the fp16 landmark; for ours it is the planes read plus the scales of the active channels, which the fidelity tables report to the nearest bit and which is exactly what the timing harness transfers.

\section{Results}
\S\ref{sec:offload} measures the target regime, a decode step with the KV cache and the scan index in host memory; \S\ref{sec:h2h} compares with SparQ $r=16$, the scan that takes the same GPU time; \S\ref{sec:fidelity} to \S\ref{sec:ruler} establish the accuracy side at equal bytes, across selection ratios and downstream.

\subsection{The target regime: decoding with the KV cache and its index in host memory}\label{sec:offload}
\begin{figure*}[t]\centering
\fig{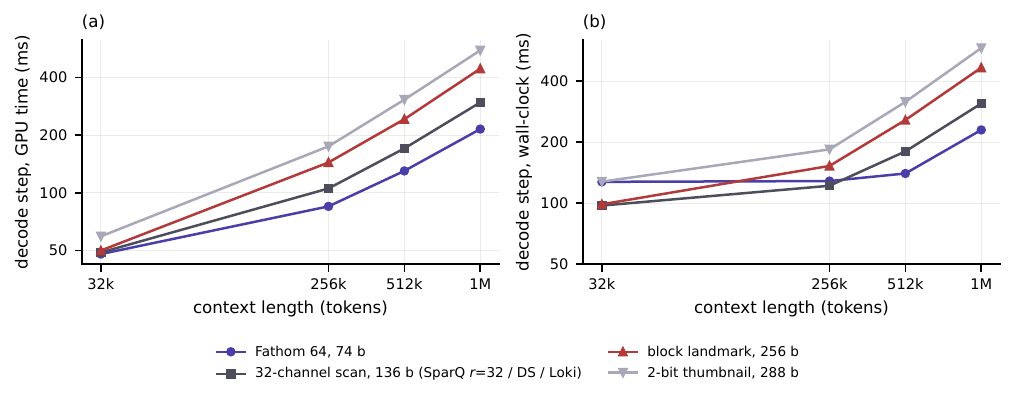}{\textwidth}{offload wall and GPU time}
\caption{Decode step versus context with K/V rows and scan index in pinned host memory (A100, Qwen3-8B, $k=512$, batch 1, synthetic KV at every point), log scale. (a) GPU time, the primary metric; (b) wall-clock. Our 74-bit read is drawn because it is the budget that reaches the error of the two 136-bit scans across the settings of Table~\ref{tab:equalerror}; the landmark index and the thumbnail are at their own cost and are not matched on that metric. Table~\ref{tab:offload} gives every budget and SparQ $r=16$, and \S\ref{sec:h2h} compares that read with time held fixed. Every method's index is moved with the contiguous gather of \S\ref{sec:gather}.}
\label{fig:offload}
\end{figure*}
\begin{table*}[t]\centering\small
\caption{Decode step (ms) on an A100 with K/V rows and the scan index in pinned host memory (Qwen3-8B, $k=512$, batch 1, synthetic KV). bits is the measured scan traffic per token; GB/step is the measured PCIe traffic at 1M. GPU time is the profiled step; wall-clock is the median of the timed steps; ratios are computed from unrounded times.}
\label{tab:offload}
\resizebox{\textwidth}{!}{\begin{tabular}{lrrrrrrrrrr}
\toprule
 & & \multicolumn{2}{c}{256k} & \multicolumn{2}{c}{512k} & \multicolumn{2}{c}{1M} & \multicolumn{3}{c}{1M relative to Fathom 48} \\
method & bits & wall & GPU & wall & GPU & wall & GPU & GB/step & wall & GPU \\
\midrule
\textbf{Fathom 40} & 47 & 129 & \textbf{73} & 129 & \textbf{106} & 224 & \textbf{168} & \textbf{1.96} & 1.00 & 0.95 \\
\textbf{Fathom 48} & 56 & 129 & 77 & 130 & 113 & 225 & 177 & 2.32 & 1.00 & 1.00 \\
\textbf{Fathom 64} & 74 & 129 & 85 & 140 & 130 & 230 & 215 & 3.00 & 1.02 & 1.21 \\
SparQ $r{=}16$ & 68 & 121 & 78 & 126 & 115 & 210 & 186 & 2.76 & 0.94 & 1.05 \\
32-channel scan & 136 & 122 & 105 & 180 & 170 & 310 & 296 & 5.33 & 1.38 & 1.67 \\
block landmark & 256 & 153 & 144 & 257 & 242 & 466 & 444 & 9.89 & 2.07 & 2.50 \\
2-bit thumbnail & 288 & 184 & 175 & 316 & 306 & 583 & 554 & 11.09 & 2.59 & 3.12 \\
\bottomrule
\end{tabular}}
\end{table*}
\begin{table*}[t]\centering\small
\caption{Real prefill, Qwen3-8B, $k=512$, batch 1, A100: GPU time per step (ms) with the scan index in host memory and with the same index kept in HBM (rows in host memory in both). The 32-channel scan is the SparQ $r=32$, Double Sparsity and Loki byte count.}
\label{tab:hbm}
\resizebox{0.7\textwidth}{!}{\begin{tabular}{lrrrrrr}
\toprule
 & \multicolumn{3}{c}{index in host memory} & \multicolumn{3}{c}{index in HBM} \\
method & 32k & 64k & 128k & 32k & 64k & 128k \\
\midrule
Fathom 40 & 47 & 49 & 56 & 43 & 44 & 47 \\
Fathom 48 & 47 & 50 & 58 & 43 & 44 & 47 \\
Fathom 64 & 48 & 52 & 62 & 44 & 45 & 48 \\
SparQ $r{=}16$ & 46 & 50 & 58 & 42 & 43 & 45 \\
32-channel scan & 50 & 57 & 73 & 43 & 44 & 47 \\
block landmark & 51 & 64 & 89 & 39 & 42 & 44 \\
2-bit thumbnail & 60 & 76 & 108 & 46 & 48 & 55 \\
dense (all rows) & 216 & 411 & 804 & 215 & 413 & 807 \\
\bottomrule
\end{tabular}}
\end{table*}

With the index in host memory the scan share of the step grows with context and the byte ratio approaches the time ratio (Table~\ref{tab:offload}, Fig.~\ref{fig:offload}). At 1M tokens the step's GPU time with our 56-bit read is \ratioGPUchanr{}$\times$ lower than with the 32-channel scan, which is the SparQ $r=32$, Double Sparsity and Loki byte count, \ratioGPUlandmark$\times$ lower than with the landmark index and \ratioGPUplanesthumb$\times$ lower than with the thumbnail; in wall-clock the ratios are \ratioWallchanr, \ratioWalllandmark{} and \ratioWallplanesthumb$\times$. SparQ $r=16$ moves \sparqSixteenBytesMorePct\% more scan bytes than our 56-bit read and takes \ratioGPUchanrsixteen$\times$ its GPU time: our scan kernel does more arithmetic per byte than a nibble scan, and at this link rate that offsets most of the transfer saving; its error is higher on every setting (\S\ref{sec:fidelity}). Our mean-40 read, at \fortyBytesFewerPct\% fewer bytes than SparQ $r=16$, is \ratioSparqSixteenOverForty$\times$ faster. In wall-clock SparQ $r=16$ is faster: the gap between wall-clock and GPU time is \issueplanesmean\,ms per step for our 56-bit read against \issuechanrsixteen\,ms for SparQ (Table~\ref{tab:offload}; A6 in \S\ref{sec:abl} discusses this gap). Our mean-64 read (74 bits) costs \ratioGPUplanesmeansixtyfour$\times$ the 56-bit read. At 256k the shared row fetch, \rowsMBmin--\rowsMBmax\,MB per step across methods, is a large part of the step and GPU time favours the 56-bit read over the 32-channel scan by \ratioGPUchanAt{}$\times$. With a real prefill at 32k--128k the ordering is the same, \realRatio{}$\times$ over the 32-channel scan at 128k in GPU time and \realRatioSparq{}$\times$ against SparQ $r=16$ (Table~\ref{tab:hbm}, left). At batch 2 with the synthetic cache the GPU-time ratio over the 32-channel scan is \bTwoRatioShort$\times$ at 256k and \bTwoRatioLong$\times$ at 512k.

\begin{table*}[t]\centering\small
\caption{Where the 1M-token step goes: GPU time per component (ms) from the profiler trace of the host-index run, Qwen3-8B, batch 1. ``scan total'' is transfer plus kernel, the part a scan method changes; top-$k$, row fetch and the weight matrix multiplications (GEMMs) are shared by every method. The landmark index scores with a GEMM, counted in the GEMM column.}
\label{tab:breakdown}
\resizebox{\textwidth}{!}{\begin{tabular}{lrrrrrrrr}
\toprule
method & scan transfer & scan kernel & top-$k$ & row fetch & GEMMs & other & scan total & step \\
\midrule
\textbf{Fathom 40} & 68 & 19 & 36 & 15 & 10 & 19 & 88 & 168 \\
\textbf{Fathom 48} & 78 & 21 & 35 & 15 & 10 & 19 & 99 & 177 \\
\textbf{Fathom 64} & 107 & 27 & 36 & 16 & 10 & 19 & 134 & 215 \\
SparQ $r{=}16$ & 96 & 11 & 36 & 15 & 10 & 19 & 106 & 186 \\
32-channel scan & 197 & 20 & 36 & 15 & 10 & 19 & 216 & 296 \\
block landmark & 359 & 0 & 7 & 9 & 29 & 40 & 359 & 444 \\
2-bit thumbnail & 406 & 67 & 35 & 17 & 10 & 19 & 473 & 554 \\
\bottomrule
\end{tabular}}
\end{table*}
Table~\ref{tab:breakdown} splits the 1M step into components. Every scan moves its bytes at the link rate, about 26\,GB/s, so the transfer column is the byte count over the link rate and is where Fathom saves; its scan kernel costs about twice SparQ's per byte because it extracts bits, and the shared costs of top-$k$ selection, the winner-row fetch and the weight GEMMs are \sharedPctOurs\% of Fathom's step and \sharedPctChan\% of the 32-channel scan's. Against SparQ $r=16$ the 56-bit read's transfer saving is mostly offset by its extra kernel time, which is the small margin in the table; the mean-40 read, at \fortyBytesFewerPct\% fewer scan bytes than SparQ $r=16$, takes $1/$\ratioSparqSixteenOverForty{} of its GPU time and $1/$\ratioChanOverForty{} of the 32-channel scan's.

\paragraph{Control: index in HBM.} If the same indices are kept in HBM and only the rows are offloaded (Table~\ref{tab:hbm}, right; Fig.~\ref{fig:hbm}), the per-token scans land at \hbmScanMin--\hbmScanMax\,ms per step at 128k, the landmark index at \hbmLandmark\,ms and the thumbnail at \hbmThumb\,ms, and ours is not faster. The time advantage therefore requires the index itself to live in the slower tier, which is the case of long contexts with many concurrent sessions, where a 17-byte-per-token label cache (5.1\,GB per million-token sequence for Qwen3-8B) or a 32-byte-per-token landmark index (9.7\,GB) does not fit next to the weights.

\subsection{Comparison at matched GPU time}\label{sec:h2h}
\begin{table*}[t]\centering\small
\caption{Fathom's 47-bit and 56-bit reads and SparQ's 68-bit read ($r=16$), the two scans that take the same GPU time per step: attention-output error on the seven headline settings. Bold marks the lower error; the ratio uses the better of the flat budget and the per-layer plan.}
\label{tab:headtohead}
\resizebox{0.85\textwidth}{!}{\begin{tabular}{llrrrrrrr}
\toprule
 & & SparQ $r{=}16$ & \multicolumn{3}{c}{Fathom, mean 40 ($\approx$47 bits)} & \multicolumn{3}{c}{Fathom, mean 48 ($\approx$56 bits)} \\
Model & ctx/$k$ & 68 bits & flat & per-layer & ratio & flat & per-layer & ratio \\
\midrule
Qwen3-8B & 16k/256 & 0.0077 & \textbf{0.0075} & \textbf{0.0033} & 2.3$\times$ & \textbf{0.0052} & \textbf{0.0021} & 3.7$\times$ \\
Qwen3-8B & 32k/512 & 0.0080 & 0.0142 & 0.0091 & 0.9$\times$ & 0.0081 & \textbf{0.0074} & 1.1$\times$ \\
Qwen3-4B & 16k/256 & 0.0071 & \textbf{0.0050} & \textbf{0.0027} & 2.7$\times$ & \textbf{0.0034} & \textbf{0.0019} & 3.8$\times$ \\
Llama-3.1-8B & 4k/128 & 0.0044 & \textbf{0.0013} & \textbf{0.0012} & 3.8$\times$ & \textbf{0.0009} & \textbf{0.0010} & 4.9$\times$ \\
Qwen2.5-7B & 32k/512 & 0.0090 & \textbf{0.0065} & 0.0096 & 1.4$\times$ & \textbf{0.0044} & \textbf{0.0063} & 2.0$\times$ \\
Qwen2.5-7B-1M & 32k/512 & 0.0091 & \textbf{0.0052} & \textbf{0.0075} & 1.7$\times$ & \textbf{0.0038} & \textbf{0.0038} & 2.4$\times$ \\
Qwen2.5-7B-1M & 128k/2048 & 0.0055 & \textbf{0.0019} & \textbf{0.0017} & 3.2$\times$ & \textbf{0.0013} & \textbf{0.0011} & 5.3$\times$ \\
\bottomrule
\end{tabular}}
\end{table*}
Among the scans in Table~\ref{tab:offload}, SparQ at $r=16$ is the one whose step time matches ours: it reads 68 bits per token to our 56, and the two take the same GPU time per step, SparQ \ratioGPUchanrsixteen$\times$ ours at 1M tokens (Table~\ref{tab:offload}) and \realRatioSparq$\times$ at 128k with a real prefill (Table~\ref{tab:hbm}). This section therefore holds time fixed and compares bytes and accuracy. At equal time Fathom reads \fortyEightBytesFewerPct\% fewer scan bytes and has \hthRatioMin--\hthRatioMax$\times$ lower attention-output error on all seven settings with the better of its two plans, and on six with the flat default (Table~\ref{tab:headtohead}). The margin is largest at the selection ratio that matters downstream (\hthRatioLong$\times$ at $k=2048$ on Qwen2.5-7B-1M at 128k) and smallest on Qwen3-8B at 32k with the flat budget, where the two are close to a tie. At the mean-40 budget, about 47 bits, Fathom's error is lower on \hthFortyWins{} of the \hthFortyN{} settings and its step is faster, SparQ $r=16$ taking \ratioSparqSixteenOverForty$\times$ its GPU time at 1M.

Two measurements qualify this. In wall-clock, SparQ $r=16$ is \wallGapSparqPct\% faster at 1M; the difference between wall-clock and GPU time is host-side overhead of this research harness (\issueplanesmean{} versus \issuechanrsixteen\,ms per step, \S\ref{sec:abl} A6), not transfer or kernel time. With every index resident in HBM the two methods take the same time (\S\ref{sec:analysis}). SparQ at $r=32$, 136 bits, is more accurate than our 56-bit read on three of the seven settings (Qwen3-8B at 32k, Qwen2.5-7B and Qwen2.5-7B-1M at 32k) and costs \ratioGPUchanr$\times$ its GPU time; the 74- to 92-bit reads reach its accuracy there, the 74-bit read at a $1.4\times$ saving (Table~\ref{tab:offload}).

\subsection{Fidelity at equal error}\label{sec:fidelity}
\begin{figure*}[t]\centering
\fig{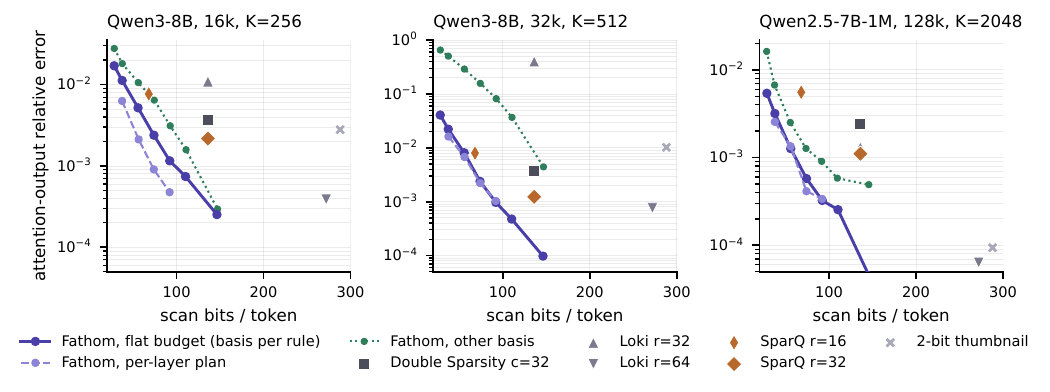}{\textwidth}{bits vs error frontier}
\caption{Attention-output relative error versus scan bits per token. Ours is the curve, flat budgets and per-layer plans, in the basis the rule of \S\ref{sec:basis} selects; baselines are points at their bit cost. (a) Qwen3-8B, 16k, $k=256$. (b) Qwen3-8B, 32k, $k=512$. (c) Qwen2.5-7B-Instruct-1M, 128k, $k=2048$.}
\label{fig:frontier}
\end{figure*}

\begin{table*}[t]\centering\small
\caption{Bits per token at which our scan reaches the error of the two 136-bit scans, on \nSettings{} settings. ``plan'' is the cheapest of the flat budget and the per-layer plan that reaches the target; basis follows the rule of \S\ref{sec:basis}.}
\label{tab:equalerror}
\resizebox{\textwidth}{!}{\begin{tabular}{llllrlllrll}
\toprule
 & & & \multicolumn{4}{c}{Double Sparsity $c{=}32$ (136 b)} & \multicolumn{4}{c}{SparQ $r{=}32$ (136 b)} \\
Model & ctx/$k$ & basis & error & Fathom bits & Fathom error & plan & error & Fathom bits & Fathom error & plan \\
\midrule
Qwen3-8B & 16k/256 & raw & 0.0037 & \textbf{47} & 0.0033 & per-layer & 0.0022 & \textbf{56} & 0.0021 & per-layer \\
Qwen3-8B & 32k/512 & raw & 0.0037 & \textbf{74} & 0.0024 & flat & 0.0012 & \textbf{92} & 0.0010 & flat \\
Qwen3-4B & 16k/256 & raw & 0.0030 & \textbf{46} & 0.0027 & per-layer & 0.0022 & \textbf{56} & 0.0019 & per-layer \\
Llama-3.1-8B & 4k/128 & KLT & 0.0015 & \textbf{47} & 0.0013 & flat & 0.0016 & \textbf{38} & 0.0015 & per-layer \\
Qwen2.5-7B & 32k/512 & KLT & 0.0043 & \textbf{74} & 0.0025 & flat & 0.0025 & \textbf{74} & 0.0025 & flat \\
Qwen2.5-7B-1M & 32k/512 & KLT & 0.0042 & \textbf{56} & 0.0038 & flat & 0.0020 & \textbf{74} & 0.0019 & flat \\
Qwen2.5-7B-1M & 128k/2048 & KLT & 0.0024 & \textbf{47} & 0.0019 & flat & 0.0011 & \textbf{56} & 0.0011 & per-layer \\
\bottomrule
\end{tabular}}
\end{table*}

Table~\ref{tab:equalerror} and Figure~\ref{fig:frontier} give the bits at which our scan reaches the error of Double Sparsity and of SparQ $r=32$, both at 136 bits. Double Sparsity's error is reached at \matchDSmin--\matchDSmax{} bits on all \nSettings{} settings, a \matchDSratioMin--\matchDSratioMax$\times$ saving; SparQ $r=32$, the strongest fixed-depth scan on six of the seven settings, is reached at \matchSQmin--\matchSQmax{} bits. SparQ $r=16$ at 68 bits has higher error than our 56-bit read on every setting, by the flat budget on \sparqSixteenFlatWins{} of them and by the per-layer plan on the rest (Tables~\ref{tab:qwen3_8b} and \ref{tab:others}). Loki at 136 bits is matched at \matchLokiMin--\matchLokiMax{} bits on the \matchLokiN{} settings where it is not rank-limited; on the Qwen3 models it collapses (\S\ref{sec:abl}). Tables~\ref{tab:qwen3_8b} and \ref{tab:others} list every method's row, with the full 4-bit scan as the floor.

\subsection{Effect of the selection ratio}\label{sec:ratio}

A 128k run at $k=128$ selects 0.1\% of the keys; the fixed-depth scans' error rises \riseSparq--\riseDS$\times$ relative to the 1.6\% ratio, ours \riseOurs$\times$ and the full 4-bit scan's far more, so the margin over SparQ $r=16$ narrows from \marginHighRatio$\times$ at 1.6\% to \marginLowRatio$\times$ at 0.1\%; at the matched ratio ($k=2048$) the crossings with the 136-bit scans are where they are at 32k (Table~\ref{tab:kratio}, Fig.~\ref{fig:ratio}). The saving does not shrink with context length; it shrinks with the selection ratio, for every scan, and at 0.1\% the rotated store lowers the 56-bit error from \kratioRawFortyEight{} to \kratioKltFortyEight.

\subsection{Downstream quality}\label{sec:ruler}
\begin{table*}[t]\centering\small
\caption{RULER-style tasks: mean score over tasks $\pm$ the standard error of the pooled per-sample scores. Bits as in \S\ref{sec:fidelity}. ``Fathom'' rows are raw planes on Qwen3-8B and ``Fathom KLT'' rows the rotated planes the basis rule selects for Qwen2.5-7B-1M. Per-task scores in Appendix~\ref{app:ruler}.}
\label{tab:ruler}
\resizebox{0.8\textwidth}{!}{\begin{tabular}{lrlrl}
\toprule
 & \multicolumn{2}{c}{Qwen3-8B 32k, $k{=}128$} & \multicolumn{2}{c}{Qwen2.5-7B-1M 128k, $k{=}128$} \\
method & bits & score & bits & score \\
\midrule
dense & -- & 0.884 $\pm$ 0.015 & -- & 0.677 $\pm$ 0.033 \\
exact top-$k$ oracle & -- & 0.874 $\pm$ 0.014 & -- & 0.660 $\pm$ 0.033 \\
\textbf{Fathom 48} & 56 & 0.881 $\pm$ 0.014 & -- & -- \\
\textbf{Fathom 64} & 74 & 0.878 $\pm$ 0.014 & -- & -- \\
\textbf{Fathom KLT 48} & -- & -- & 56 & 0.671 $\pm$ 0.032 \\
\textbf{Fathom KLT 64} & -- & -- & 74 & 0.676 $\pm$ 0.035 \\
Double Sparsity $c{=}32$ & 136 & 0.880 $\pm$ 0.014 & 136 & 0.655 $\pm$ 0.034 \\
SparQ $r{=}16$ & 68 & 0.882 $\pm$ 0.014 & 68 & 0.644 $\pm$ 0.034 \\
SparQ $r{=}32$ & 136 & 0.881 $\pm$ 0.014 & 136 & 0.640 $\pm$ 0.034 \\
Loki $r{=}64$ & 272 & 0.882 $\pm$ 0.014 & 272 & 0.655 $\pm$ 0.033 \\
2-bit thumbnail & 288 & 0.877 $\pm$ 0.014 & 288 & 0.685 $\pm$ 0.035 \\
block landmark & 256 & 0.831 $\pm$ 0.016 & 256 & 0.531 $\pm$ 0.038 \\
\bottomrule
\end{tabular}}
\end{table*}

With \rulerN{} samples per task at 32k and \rulerLongN{} at 128k, the standard error of a method's mean score is about \rulerSEShort{} and \rulerSELong. Every per-token scan lies within \rulerScanDevShort{} of the exact top-$k$ oracle at 32k and within \rulerScanDevLong{} at 128k, inside that error, and ours does so at 56 and 74 bits (Table~\ref{tab:ruler}, Fig.~\ref{fig:ruler}). The block-landmark index loses \rulerLandmarkGapShort{} against the oracle at 32k and \rulerLandmarkGapLong{} at 128k, because when $k$ is small relative to the block a block's mean hides its best token. No per-token scan separates from another at this $k$; differences between them appear only in attention error and in bytes. These tasks validate the scan as a retrieval mechanism over long contexts; task success on real coding or tool-use sessions is not measured; \S\ref{sec:agent} measures step agreement on such sessions.

\subsection{Real agent sessions}\label{sec:agent}
\begin{table*}[t]\centering\small
\caption{Long coding-agent sessions: \agentN{} sessions of about \agentCtxMean{}k tokens built from real OpenHands trajectories on one repository each; every method greedily decodes the agent's next step from the same cache (Qwen2.5-7B-Instruct-1M, $k=512$). Step agreement is the word-level sequence-match ratio between a method's step and the exact top-$k$ step, with its standard error over sessions. Bold marks Fathom.}
\label{tab:agent}
\resizebox{0.55\textwidth}{!}{\begin{tabular}{lrl}
\toprule
method & bits & step agreement with exact top-$k$, mean $\pm$ s.e. \\
\midrule
exact top-$k$ oracle & -- & 1.00 $\pm$ 0.00 \\
\textbf{Fathom 40} & 47 & \textbf{0.49} $\pm$ 0.04 \\
\textbf{Fathom 48} & 56 & \textbf{0.53} $\pm$ 0.04 \\
\textbf{Fathom 64} & 74 & \textbf{0.54} $\pm$ 0.04 \\
\textbf{Fathom 80} & 92 & \textbf{0.60} $\pm$ 0.05 \\
Double Sparsity $c{=}32$ & 136 & 0.55 $\pm$ 0.04 \\
SparQ $r{=}16$ & 68 & 0.49 $\pm$ 0.04 \\
SparQ $r{=}32$ & 136 & 0.60 $\pm$ 0.04 \\
block landmark & 256 & 0.42 $\pm$ 0.03 \\
\bottomrule
\end{tabular}}
\end{table*}
RULER-style tasks share structure with agent sessions, not content. Table~\ref{tab:agent} therefore decodes real ones: trajectories of the OpenHands agent~\cite{openhands} on SWE-rebench issues~\cite{swerebench}, taken as recorded (system prompt, task, tool calls, file contents and command output), concatenated per repository until the session holds 80k--100k tokens, with the agent's next recorded step as the target. Each method decodes that step from the same prefill. The baseline is exact top-$k$ decoding with the same $k$, which is what every scan is built to reproduce, and the score is step agreement: the word-level sequence-match ratio between a method's step and the exact top-$k$ step.

At $k=2048$ (2\% of the context) Fathom at 56 bits agrees with the exact top-$k$ step at \agentOsimBigplanesK{} against \agentOsimBigsparq{} for SparQ $r=16$ and \agentOsimBiglandmark{} for the block landmark index. Paired per session the margin over SparQ $r=16$ is \agentPairedBigSparq{} $\pm$ \agentPairedSEBigSparq{} (\agentWinsBigSparq{} of \agentNBig{} sessions), at \fortyEightBytesFewerPct\% fewer bytes and the same GPU time. At $k=512$ (0.5\% of the context) the most accurate scan is SparQ $r=32$ at 136 bits, \agentOsimsparqthirtytwo{} (Table~\ref{tab:agent}). Fathom reaches the same agreement at 92 bits, \agentOsimplanesKeighty{} (paired \agentPairedEightySparqThirtyTwo{} $\pm$ \agentPairedSEEightySparqThirtyTwo{}, \agentWinsEightySparqThirtyTwo{} of \agentN{} sessions), 32\% fewer bytes; its 56- and 74-bit reads sit with Double Sparsity's 136-bit scan and at or above SparQ $r=16$ (\agentOsimplanesK{} against \agentOsimsparq{}, paired \agentPairedSparq{} $\pm$ \agentPairedSESparq), and the landmark index is below all of them. This is what the attention-error tables predict at this selection ratio: SparQ $r=32$'s error falls between Fathom's 74- and 92-bit reads at 0.4\% and between its 56- and 74-bit reads at 1.6\% (Table~\ref{tab:kratio}), so the budget at which Fathom matches it, and the byte saving that follows, move with $k/n$ (\matchSQmin--\matchSQmax{} bits across settings, \S\ref{sec:fidelity}). Fathom's channel statistics are calibrated once on Wikitext; recalibrating them on held-out agent transcripts, or on the session's own prefill keys with no offline data at all, raises step agreement by \agentPairedAgentFortyEightWiki{} to \agentPairedAgentEightyWiki{} (within 1.5 standard errors) and changes attention error on agent-transcript keys by under 10\%, so the ordering against SparQ $r=32$ is set by the byte budget, not by the calibration domain (Appendix~\ref{app:abl}, B5). The run was planned at 100 sessions and stopped at \agentN{} for budget; the $k=2048$ subset has \agentNBig. End-to-end task success with test execution under each scan is not measured and is left to future work.
\begin{table}[t]\centering\small
\caption{The same sessions at $k=2048$ (2\% of the context), \agentNBig{} sessions: with the larger budget the scans separate.}
\label{tab:agent_kbig}
\resizebox{\columnwidth}{!}{\begin{tabular}{lrl}
\toprule
method & bits & step agreement with exact top-$k$, mean $\pm$ s.e. \\
\midrule
exact top-$k$ oracle & -- & 1.00 $\pm$ 0.00 \\
\textbf{Fathom 48} & 56 & \textbf{0.67} $\pm$ 0.06 \\
SparQ $r{=}16$ & 68 & 0.49 $\pm$ 0.05 \\
block landmark & 256 & 0.47 $\pm$ 0.05 \\
\bottomrule
\end{tabular}}
\end{table}

\section{Ablations}\label{sec:abl}
Each ablation changes one design choice and holds the rest fixed. Unless noted, the metric is held-out attention-output error at matched scan bits.

\paragraph{A1. Per-layer versus flat budgets.} Table~\ref{tab:plan} compares the flat budget with the greedy per-layer plan at means 48 and 64 on every setting. The plan helps on the Qwen3 models at 16k and less at 32k, hurts on Qwen2.5-7B and Llama-3.1-8B, and is neutral to better on Qwen2.5-7B-1M; a 16k plan evaluated at 32k on Qwen3-8B is worse than the flat budget (the ``plan from 16k'' rows of Table~\ref{tab:qwen3_8b}). Flat is the default. Figure~\ref{fig:layers} shows the plans.
\begin{table}[t]\centering\small
\caption{A1: flat budget versus per-layer plan, error at means 48 and 64, in the basis the rule selects.}
\label{tab:plan}
\resizebox{\columnwidth}{!}{\begin{tabular}{lllllll}
\toprule
 & & & \multicolumn{2}{c}{mean 48} & \multicolumn{2}{c}{mean 64} \\
Model & ctx/$k$ & basis & flat & per-layer & flat & per-layer \\
\midrule
Qwen3-8B & 16k/256 & raw & 0.0052 & 0.0021 & 0.0024 & 0.0010 \\
Qwen3-8B & 32k/512 & raw & 0.0081 & 0.0074 & 0.0024 & 0.0021 \\
Qwen3-8B & 32k/128 & raw & 0.0393 & 0.0363 & 0.0170 & 0.0182 \\
Qwen3-4B & 16k/256 & raw & 0.0034 & 0.0019 & 0.0021 & 0.0007 \\
Llama-3.1-8B & 4k/128 & KLT & 0.0009 & 0.0010 & 0.0004 & 0.0009 \\
Qwen2.5-7B & 32k/512 & KLT & 0.0044 & 0.0063 & 0.0025 & 0.0044 \\
Qwen2.5-7B & 32k/128 & KLT & 0.0215 & 0.0236 & 0.0127 & 0.0141 \\
Qwen2.5-7B-1M & 32k/512 & KLT & 0.0038 & 0.0038 & 0.0019 & 0.0019 \\
Qwen2.5-7B-1M & 128k/128 & KLT & 0.0196 & 0.0185 & 0.0111 & 0.0103 \\
Qwen2.5-7B-1M & 128k/512 & KLT & 0.0048 & 0.0040 & 0.0025 & 0.0017 \\
Qwen2.5-7B-1M & 128k/2048 & KLT & 0.0013 & 0.0011 & 0.0006 & 0.0004 \\
\bottomrule
\end{tabular}}
\end{table}

\paragraph{A2. Basis.} Table~\ref{tab:basis} and Figure~\ref{fig:basis} compare raw and KLT-rotated planes at the same budgets on every setting. The rotation raises error on Qwen3 and lowers it on the three models without QK-norm. QK-norm equalizes channel scales and leaves the per-query sparsity in the raw basis; without it, the rotation concentrates it.
\begin{table}[t]\centering\small
\caption{A2: raw versus KLT-rotated planes, error at means 48 and 64.}
\label{tab:basis}
\resizebox{\columnwidth}{!}{\begin{tabular}{llllll}
\toprule
 & & \multicolumn{2}{c}{mean 48} & \multicolumn{2}{c}{mean 64} \\
Model & ctx/$k$ & raw & KLT & raw & KLT \\
\midrule
Qwen3-8B & 16k/256 & 0.0052 & 0.0106 & 0.0024 & 0.0064 \\
Qwen3-8B & 32k/512 & 0.0081 & 0.2908 & 0.0024 & 0.1576 \\
Qwen3-4B & 16k/256 & 0.0034 & 0.0167 & 0.0021 & 0.0102 \\
Llama-3.1-8B & 4k/128 & 0.0025 & 0.0009 & 0.0012 & 0.0004 \\
Qwen2.5-7B & 32k/512 & 0.0075 & 0.0044 & 0.0043 & 0.0025 \\
Qwen2.5-7B & 32k/128 & 0.0264 & 0.0215 & 0.0153 & 0.0127 \\
Qwen2.5-7B-1M & 32k/512 & 0.0066 & 0.0038 & 0.0038 & 0.0019 \\
Qwen2.5-7B-1M & 128k/128 & 0.0300 & 0.0196 & 0.0162 & 0.0111 \\
Qwen2.5-7B-1M & 128k/512 & 0.0082 & 0.0048 & 0.0036 & 0.0025 \\
Qwen2.5-7B-1M & 128k/2048 & 0.0025 & 0.0013 & 0.0013 & 0.0006 \\
\bottomrule
\end{tabular}}
\end{table}

\paragraph{A3. Selection ratio.} In Table~\ref{tab:kratio} the crossing with the 136-bit scans moves to higher budgets as $k/n$ falls from 1.6\% to 0.1\%, with raw and with rotated planes. The effect is common to every scan including the full 4-bit one.

\paragraph{A4. Uniform versus water-filled depth.} The 2-bit thumbnail reads every channel at two planes, 288 bits per token. Its error in Tables~\ref{tab:qwen3_8b} and \ref{tab:others} is reached by the water-filled read at \thumbMatchMin--\thumbMatchMax{} bits on all \nSettings{} settings (the 148-bit case is Qwen2.5-7B-1M at 128k with $k=2048$, where the thumbnail is also strong). Uniform depth spends bits on channels that carry no score for the query.

\paragraph{A5. SparQ's selection rule.} SparQ's grouped-query rule reads one channel set per KV head. Letting every query head pick its own $r$ channels and reading the union lowers error but costs $2.1$--$3\times$ the bytes (Table~\ref{tab:sparqvariant}, Appendix~\ref{app:abl}). The $r=16$ variant sits at 148--200 bits and has higher error than our 146-bit read on every setting (Tables~\ref{tab:qwen3_8b} and \ref{tab:others}); the $r=32$ variant reads 284--373 bits, beyond any budget we measure.

\paragraph{A6. Measurement controls.} Wall-clock exceeds GPU time by the time the GPU waits on the Python harness. At 1M the gap is \issueplanesmean\,ms per step for our 56-bit read, \issueplanesmeansixtyfour\,ms for the 74-bit read, \issuechanrsixteen{} and \issuechanr\,ms for the 16- and 32-channel scans, \issuelandmark\,ms for the landmark index and \issueplanesthumb\,ms for the thumbnail (Table~\ref{tab:offload}). The two plane reads launch the same kernels yet show different gaps, which we did not isolate; the gap is host-side and absent from GPU time, and a fused implementation would not have it, which is why GPU time is the primary metric. The wall-clock ratios of every baseline are below the GPU-time ratios. Every timing row uses a fixed-size run list with no host synchronization, and the gather's staging buffer was checked bit-exact against a direct GPU scan over the same keys on this pod before any timing was recorded. Two engineering variants were tried and are not used: a chunked pipeline that overlaps the gather of chunk $i{+}1$ with the scan of chunk $i$ on a second stream, bit-exact but no faster, because both gathers already run at the link rate and the scan kernel slows under contention; and a sweep of the Triton launch configurations of both scan kernels and a tensor-core variant, none faster than the configurations reported. Two-stage and reduced-precision top-$k$ were no faster than the library top-$k$ either.

\section{Analysis}\label{sec:analysis}
\paragraph{HBM-resident scan kernel.}\label{sec:kernel}

With everything in HBM the byte saving does not become time (Table~\ref{tab:kernel}, Fig.~\ref{fig:kernel}). The planes kernel extracts one bit per shift-and-mask and applies $G$ multiply-adds per extracted bit, where a 4-bit nibble carries four times the information per extracted element, so it reaches a fraction of the bandwidth the nibble scans reach. The 32-channel scan's kernel time is \kernelRatioShort$\times$ ours at 32k and \kernelRatioLong$\times$ at 128k; ours is slower in the kernel despite reading \kernelBytesFewerPct\% fewer bytes on these layers. The 32-channel scan is \scanShareShort--\scanShareLong\% of its sparse-attention step and ours \scanShareOursShort--\scanShareOursLong\% (top-$k$, gather and exact attention are method-independent), so at the step level the per-token scans land within 6\% of each other (\hbmScanMin--\hbmScanMax\,ms at 128k) and ours is not the fastest, which is what the HBM-index columns of Table~\ref{tab:hbm} show.

\paragraph{Arithmetic per byte.} The scan's arithmetic is one multiply-add per (active channel, token, head), so reading fewer planes of a channel removes bytes, not multiply-adds. Bit extraction adds one to two integer operations per bit read, four times the per-bit cost of a nibble scan. Counting INT32 lanes times clock against HBM bandwidth (108 SMs, 64 lanes, 1.41\,GHz, 2.0\,TB/s), an A100 offers about 5 integer operations per byte, an H100 SXM about the same and a bandwidth-poor L4 about 25, so a scan that is arithmetic-bound on the A100 is arithmetic-bound on any current data-centre GPU; the HBM-resident behaviour above is a property of the operation, not of the card. Amdahl's law compounds it: where the scan is a minority of sparse attention, even a zero-time scan yields little. Over PCIe the ordering reverses. Over PCIe every method runs at the link's rate, the arithmetic is idle, and the byte ratio approaches the time ratio (Table~\ref{tab:offload}).

\paragraph{Access pattern.} The Triton gather of contiguous runs from pinned host memory moves \pcieGatherMin--\pcieGatherMax\,GB/s with the best block size at every run size from 4\,KB (one plane of one channel at 32k) to 512\,KB (four planes at 1M), and \pcieGatherWorst\,GB/s at 4\,KB with the worst. That is the rate of one \texttt{cudaMemcpyAsync} of the same bytes (\pcieMemcpyMin--\pcieMemcpyMax\,GB/s), and runs placed at $4\times$ stride lose nothing (Fig.~\ref{fig:pcie}). One copy call per run reaches \pciePercallMin\,GB/s at 4\,KB runs, about 11\,$\mu$s per call, and \pciePercallMax\,GB/s only at 512\,KB. The bit-plane layout is what makes a per-query selection of planes a set of long contiguous runs, and the gather kernel moves those runs at link rate even at 4\,KB granularity; a landmark or label-cache index is one contiguous block and needs no gather, which is why every baseline in Table~\ref{tab:offload} is given the same transfer and the comparison is one of bytes.

\paragraph{Memory.} The store has a memory cost. As a separate index it is 68 bytes per token per KV head, four times Double Sparsity's 17 and twice a 32-byte landmark index; its advantage is in bytes read per query, not bytes stored. The store becomes free only in a stack whose K cache is already 4-bit and channel-major, where the winners' keys are reconstructed from all four planes; we have not built or measured that path, and reconstructing single tokens from a channel-major layout is not cheap.

\paragraph{Sensitivity of the downstream tasks.} On RULER-style tasks at $k\ge 128$ the top-$k$ set is robust to scan error of the magnitude all per-token scans exhibit; those tasks separate selection granularity (blocks versus tokens) and the top-$k$ budget itself, not the scans. Real agent steps are more sensitive: at a 2\% budget the scan's fidelity to the exact top-$k$ step is visible in what the agent writes (\S\ref{sec:agent}). A task that separated 0.002 from 0.004 attention error would need a much smaller $k$ or a much longer dependency chain than RULER's.

\section{Limitations}\label{sec:limits}
The systems regime this paper targets, offloaded caches and million-token contexts with many concurrent sessions, is measured directly. Quality is measured on attention-output error, on synthetic retrieval and tracking tasks, and on real agent sessions scored by step agreement with exact top-$k$ decoding (\agentN{} of 100 planned sessions at $k=512$, \agentNBig{} at $k=2048$); end-to-end task success with test execution under each scan is not measured and is left to future work. With the scan index resident in HBM the method is not faster than the 4-bit channel scans; its time advantage requires the index to live in a slower tier. The landmark baseline is a reimplementation of ShadowKV's chunk means in a setting (host-resident index, the paper's $k$) that ShadowKV and Quest do not use, and CPU-side retrieval of the RetroInfer and MagicPIG kind is not compared. Timing beyond 128k used synthetic KV contents. The Triton kernels are prototypes and the offload harness pays a per-method Python issue cost that a fused implementation would not; GPU time is reported for that reason. Per-layer plans must be calibrated at the deployment context length or replaced by a flat budget. Channel statistics are calibrated offline; on agent transcripts the domain of that calibration moves step agreement by at most \agentPairedAgentEightyWiki{} and the session-calibrated variant removes the dependence (Appendix~\ref{app:abl}, B5), but domains further from both were not tested. Fidelity is measured on one held-out window per setting, and RULER-style scores carry standard errors of a few points, so differences between per-token scans smaller than that are not established. The offload tables are batch 1, with one batch-2 check; the many-session case that motivates the regime is argued from index sizes rather than measured. The instruct model at 128k was prompted without its chat template, which lowers its absolute scores without affecting the comparison between scans. Top-$k$ is selected per query head for every method; SparQ's group-shared top-$k$, which would reduce the winner rows for SparQ and for any other method, and its mean-value reallocation, which would lower SparQ's absolute error without changing the comparison of scans, are not evaluated.

\section{Conclusion}
\ours{} is a key scan for sparse decoding when the KV cache and its index live in host memory. There, a decode step at one million tokens runs \ratioGPUchanr$\times$ faster in GPU time than with the 136-bit scans and \ratioGPUlandmark$\times$ faster than with a landmark index, and against SparQ's 68-bit read it moves \fortyEightBytesFewerPct\% fewer bytes at the same GPU time with \hthRatioMin--\hthRatioMax$\times$ lower attention error, and at 47 bits it is \ratioSparqSixteenOverForty$\times$ faster. The mechanism is a 4-bit K cache stored in channel-major bit planes, so a prefix read is an exact lower-precision quantizer, and a per-query water-filling rule that allocates bits to the channels that carry the query's score variance. At Double Sparsity's error it reads \matchDSmin--\matchDSmax{} bits where fixed-depth scans read 136, on seven settings up to 128k; it matches the exact top-$k$ oracle on RULER-style tasks and, on real coding-agent sessions at a 2\% budget, agrees with the exact top-$k$ step at \agentOsimBigplanesK{} against \agentOsimBigsparq{} for SparQ $r=16$. Where everything sits in HBM the method is not faster, because the scan is arithmetic-bound there and reading fewer bits does not remove multiply-adds; a fused kernel would narrow the gap but not remove the arithmetic bound. The method fits serving stacks that already keep a 4-bit K cache and hold it in a slower tier.

\FloatBarrier
\bibliographystyle{plain}
\bibliography{refs}

\appendix
\section{Worked example}\label{app:toy}
Six keys, four channels, two query heads sharing the KV head. Keys (rows $t_0\ldots t_5$) and their channel variances:
\begin{equation*}\small
K=\begin{pmatrix}1.8&0.4&-0.2&0.05\\-1.1&0.9&0.3&-0.02\\0.3&-1.2&0.1&0.04\\1.2&0.7&0.4&0.01\\-0.6&-0.3&-0.5&0.03\\0.9&1.1&-0.3&-0.05\end{pmatrix},\quad
\mathrm{Var}=\begin{pmatrix}1.222\\0.755\\0.127\\0.001\end{pmatrix}^{\!\top}.
\end{equation*}
The queries are $(3,1,2,0.1)$ for head A and $(1,2,-1,0.2)$ for head B. Exact scores give top-2 sets $\{t_0,t_3\}$ for A and $\{t_5,t_0\}$ for B; B's decision between $t_0$ and $t_3$ rests on channel 2, where $t_3$ has $0.4$ and $t_0$ has $-0.2$.

\paragraph{Importance weights.} Eq.~\ref{eq:g} sums the squared query weights of both heads against the channel variances: $g_0 = (3^2+1^2)\,1.222 = 12.22$, $g_1 = (1^2+2^2)\,0.755 = 3.77$, $g_2 = (2^2+1^2)\,0.127 = 0.63$, $g_3 = (0.1^2+0.2^2)\,0.001 \approx 0$.

\paragraph{Finding the water line.} $\theta$ is a free parameter. For any value of $\theta$, Eq.~\ref{eq:wf} turns the four importances into four depths, and the depths have a sum. A small $\theta$ makes every $g_j/\theta$ large and hands out many bits; a large $\theta$ hands out few. The water line is the smallest $\theta$ whose depths still fit the budget $B$, because fewer bits than the budget would waste budget and more would exceed it. Bisection finds it by trial. Take $B=8$ and start from the bracket $[0.001, 10]$ on $\theta$. Each trial is the midpoint of the current bracket on a log scale, that is the geometric mean of its two ends ($\sqrt{0.001\times10}=0.1$, then $\sqrt{0.001\times0.1}=0.01$, and so on); if the trial's depths overflow $B$ it becomes the new lower end lo, otherwise the new upper end hi:
\begin{center}\setlength{\tabcolsep}{3pt}\resizebox{\columnwidth}{!}{\begin{tabular}{crcccl}
\toprule
$[\mathrm{lo},\mathrm{hi}]$ & trial $\theta$ & $\log_4(g_j/\theta)$ & $t$ & sum & \\
\midrule
$[0.001,10]$ & $0.1$ & $(3.47,2.62,1.33,-5.5)$ & $(3,3,1,0)$ & 7 & fits, hi $\leftarrow 0.1$ \\
$[0.001,0.1]$ & $0.01$ & $(5.13,4.28,2.99,-3.8)$ & $(4,4,3,0)$ & 11 & over, lo $\leftarrow 0.01$ \\
$[0.01,0.1]$ & $0.0316$ & $(4.30,3.45,2.16,-4.7)$ & $(4,3,2,0)$ & 9 & over, lo $\leftarrow 0.0316$ \\
$[0.0316,0.1]$ & $0.0562$ & $(3.88,3.03,1.74,-5.1)$ & $(4,3,2,0)$ & 9 & over, lo $\leftarrow 0.0562$ \\
$[0.0562,0.1]$ & $0.075$ & $(3.67,2.83,1.54,-5.3)$ & $(4,3,2,0)$ & 9 & over, lo $\leftarrow 0.075$ \\
$[0.075,0.1]$ & $0.0866$ & $(3.57,2.72,1.43,-5.4)$ & $(4,3,1,0)$ & 8 & fits, hi $\leftarrow 0.0866$ \\
$[0.075,0.0866]$ & $0.0806$ & $(3.62,2.77,1.48,-5.3)$ & $(4,3,1,0)$ & 8 & fits, hi $\leftarrow 0.0806$ \\
$[0.075,0.0806]$ & $0.0777$ & $(3.65,2.80,1.51,-5.3)$ & $(4,3,2,0)$ & 9 & over, lo $\leftarrow 0.0777$ \\
\bottomrule
\end{tabular}}
\end{center}
After 30 halvings the upper end has settled at $\theta = 0.079$, and the depths there, $t=(4,3,1,0)$, use all 8 bits. Channel 2 receives one plane while $\mathrm{round}(\log_4(0.63/\theta))=1$, that is while $0.5\le\log_4(0.63/\theta)<1.5$, which is $0.63/4^{1.5}<\theta\le0.63/4^{0.5}$, or $0.0788<\theta\le0.315$; just below $0.0788$ channel 2 would take a second plane and the sum would become 9. So the water line for $B=8$ sits right at the point where the next bit anywhere would overflow the budget. The same procedure for $B=6$ stops at $\theta=0.118$, where $\log_4(g_j/\theta)=(3.35, 2.50, 1.21, -5.3)$ and $t=(3,2,1,0)$; below $0.1178=3.77/4^{2.5}$ channel 1 would take a third plane. Channel 3 is never read at either budget; channel 2 gets one plane at both because its importance is a factor of six below channel 1's, and one plane costs one bit.

\paragraph{One plane of channel 2.} The block absmax of channel 2 is $a_2 = 0.5$, so the cell width is $s = a_2/8 = 0.0625$ and the 4-bit codes of the six keys are $(4,12,9,14,0,3)$, in binary $0100, 1100, 1001, 1110, 0000, 0011$. The first plane is the sign bit. Reading it alone, $c^{(1)} = c \gg 3 \in \{0,1\}$, and the dequantized value $(c^{(1)} + \tfrac12 - 1)\,a_2$ is $-0.25$ for $t_0, t_4, t_5$ and $+0.25$ for $t_1, t_2, t_3$; the true values are $-0.2, 0.3, 0.1, 0.4, -0.5, -0.3$. One bit already places $t_3$ above $t_0$ on this channel. With depths $(4,3,1,0)$ head B's approximate scores are $(2.84, 0.24, -2.01, 2.49, -0.61, 3.36)$ against exact $(2.81, 0.40, -2.19, 2.20, -0.69, 3.39)$, so B selects $\{t_5, t_0\}$ correctly; head A's are $(5.01, -1.79, 0.46, 4.96, -2.34, 3.59)$ against $(5.40, -1.80, -0.10, 5.10, -3.10, 3.19)$ and A selects $\{t_0, t_3\}$.

\begin{figure}[tbp]\centering
\fig{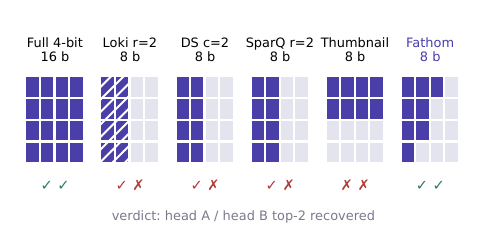}{\linewidth}{toy read shapes}
\caption{Bit-cells read per token by each method on the worked example (columns: channels, rows: planes). Checks mark heads ranked correctly.}
\label{fig:toy}
\end{figure}

\paragraph{The other methods at 8 bits.} Loki with $r=2$ keeps directions that are almost exactly channels 0 and 1 and ranks B as $\{t_5,t_3\}$. Double Sparsity with $c=2$ picks channels 0 and 1 offline and makes the same mistake. SparQ with $r=2$ under its grouped-query rule sums $|q|$ over the two heads, $(4,3,3,0.3)$, picks channels 0 and 1 (channel 1 wins the tie with channel 2 by index), reads 8 bits and ranks B as $\{t_5,t_3\}$. A 2-bit thumbnail (8 bits) reads all four channels at two planes and mis-ranks both heads. At $B=6$ Fathom's depths $(3,2,1,0)$ also mis-rank B, which is the budget at which the method fails on this example. Figure~\ref{fig:toy} draws the read shapes.

\Needspace*{10\baselineskip}
\section{RULER-style task details}\label{app:ruler}
The haystack is Wikitext-103 test text tokenized once; each sample takes a random window and inserts needles at random depths in $[0.05,0.95]$ of the window. Needle templates follow RULER, for example ``One of the special magic numbers for \{key\} is: \{7 digits\}.'' with one target key (single), one target plus seven decoy keys (multi-key; RULER's multikey-1 default is three decoys), four values for one key (multi-value), or four keys each with one value (multi-query). Variable tracking chains four assignments through one decoy chain; frequent-word extraction asks for the three most frequent synthetic words drawn with Zipf exponent 2. Generation lengths are 12 tokens for single numbers, 40 for lists and 24 for words; the score is the fraction of gold strings contained in the greedy generation. Prompts are plain completions without the chat template.
\begin{table*}[tbp]\centering\small
\caption{Qwen3-8B, 32k, $k=128$, \rulerN{} samples per task.}
\resizebox{0.7\textwidth}{!}{\begin{tabular}{lrrrrrr}
\toprule
method & niah\_single & niah\_multikey & niah\_multivalue & niah\_multiquery & vt & fwe \\
\midrule
dense & 1.000 & 0.975 & 1.000 & 0.950 & 0.485 & 0.892 \\
exact top-$k$ oracle & 1.000 & 0.975 & 1.000 & 0.938 & 0.525 & 0.808 \\
Fathom 48 & 1.000 & 0.975 & 1.000 & 0.931 & 0.565 & 0.817 \\
Fathom 64 & 1.000 & 0.975 & 1.000 & 0.931 & 0.535 & 0.825 \\
Double Sparsity $c{=}32$ & 1.000 & 0.975 & 1.000 & 0.944 & 0.560 & 0.800 \\
SparQ $r{=}16$ & 1.000 & 0.975 & 1.000 & 0.944 & 0.575 & 0.800 \\
SparQ $r{=}32$ & 1.000 & 0.975 & 1.000 & 0.938 & 0.555 & 0.817 \\
Loki $r{=}64$ & 1.000 & 0.975 & 1.000 & 0.938 & 0.545 & 0.833 \\
2-bit thumbnail & 1.000 & 0.975 & 1.000 & 0.938 & 0.535 & 0.817 \\
block landmark & 0.975 & 0.925 & 0.925 & 0.850 & 0.505 & 0.808 \\
\bottomrule
\end{tabular}}
\end{table*}
\begin{table}[tbp]\centering\small
\caption{Qwen2.5-7B-Instruct-1M, 128k, $k=128$, \rulerLongN{} samples per task.}
\resizebox{\columnwidth}{!}{\begin{tabular}{lrrrr}
\toprule
method & niah\_multikey & niah\_multiquery & vt & fwe \\
\midrule
dense & 1.000 & 0.375 & 0.550 & 0.783 \\
exact top-$k$ oracle & 1.000 & 0.388 & 0.520 & 0.733 \\
Fathom KLT 48 & 1.000 & 0.375 & 0.560 & 0.750 \\
Fathom KLT 64 & 1.000 & 0.400 & 0.520 & 0.783 \\
Double Sparsity $c{=}32$ & 1.000 & 0.375 & 0.530 & 0.717 \\
SparQ $r{=}16$ & 1.000 & 0.375 & 0.500 & 0.700 \\
SparQ $r{=}32$ & 1.000 & 0.400 & 0.510 & 0.650 \\
Loki $r{=}64$ & 1.000 & 0.400 & 0.520 & 0.700 \\
2-bit thumbnail & 1.000 & 0.400 & 0.540 & 0.800 \\
block landmark & 0.700 & 0.263 & 0.460 & 0.700 \\
\bottomrule
\end{tabular}}
\end{table}

\Needspace*{10\baselineskip}
\section{Additional tables and figures}\label{app:extra}
The tables and figures referenced from the main text but not essential to its argument.

\begin{figure*}[tbp]\centering
\fig{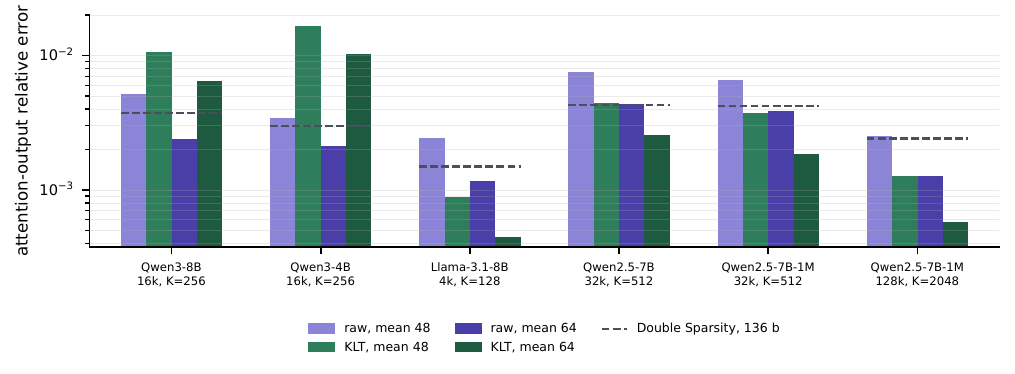}{0.8\textwidth}{basis rule}
\caption{Raw versus rotated planes at means 48 and 64 on six settings (Table~\ref{tab:basis} has all of them). Ticks mark Double Sparsity at 136 bits.}
\label{fig:basis}
\end{figure*}

\begin{table*}[tbp]\centering\small
\caption{Qwen3-8B, all methods, bits and attention-output error. Raw planes; flat budgets and per-layer plans, including the 16k plan evaluated at 32k.}
\label{tab:qwen3_8b}
\resizebox{0.62\textwidth}{!}{\begin{tabular}{lrlrl}
\toprule
 & \multicolumn{2}{c}{Qwen3-8B 16k/$K{=}256$} & \multicolumn{2}{c}{Qwen3-8B 32k/$K{=}512$} \\
 & bits & error & bits & error \\
\midrule
\textbf{Fathom flat 40} & 47 & 0.0075 & 46 & 0.0142 \\
\textbf{Fathom flat 48} & 56 & 0.0052 & 56 & 0.0081 \\
\textbf{Fathom flat 64} & 74 & 0.0024 & 74 & 0.0024 \\
\textbf{Fathom flat 80} & 92 & 0.0012 & 92 & 0.0010 \\
\textbf{Fathom flat 128} & 146 & 0.0003 & 146 & 0.0001 \\
\textbf{Fathom per-layer 40} & 47 & 0.0033 & 46 & 0.0091 \\
\textbf{Fathom per-layer 48} & 56 & 0.0021 & 55 & 0.0074 \\
\textbf{Fathom per-layer 64} & 74 & 0.0010 & 74 & 0.0021 \\
\textbf{Fathom per-layer 48, plan from 16k} & -- & -- & 56 & 0.0103 \\
\textbf{Fathom per-layer 64, plan from 16k} & -- & -- & 74 & 0.0035 \\
SparQ $r{=}16$ & 68 & 0.0077 & 68 & 0.0080 \\
SparQ $r{=}32$ & 136 & 0.0022 & 136 & 0.0012 \\
Double Sparsity $c{=}32$ & 136 & 0.0037 & 136 & 0.0037 \\
Loki $r{=}32$ & 136 & 0.0109 & 136 & 0.4033 \\
Loki $r{=}64$ & 272 & 0.0004 & 272 & 0.0008 \\
2-bit thumbnail & 288 & 0.0028 & 288 & 0.0102 \\
full 4-bit scan & 544 & $<$0.0001 & 544 & $<$0.0001 \\
\bottomrule
\end{tabular}}
\end{table*}

\begin{table*}[tbp]\centering\small
\caption{The other four headline settings. ``Fathom flat'' rows use raw planes, ``Fathom KLT'' rows the rotated planes the basis rule selects for these models.}
\label{tab:others}
\resizebox{\textwidth}{!}{\begin{tabular}{lrlrlrlrl}
\toprule
 & \multicolumn{2}{c}{Qwen3-4B 16k/$K{=}256$} & \multicolumn{2}{c}{Llama-3.1-8B 4k/$K{=}128$} & \multicolumn{2}{c}{Qwen2.5-7B 32k/$K{=}512$} & \multicolumn{2}{c}{Qwen2.5-7B-1M 32k/$K{=}512$} \\
 & bits & error & bits & error & bits & error & bits & error \\
\midrule
\textbf{Fathom flat 40} & 46 & 0.0050 & 48 & 0.0033 & 46 & 0.0107 & 47 & 0.0098 \\
\textbf{Fathom flat 48} & 56 & 0.0034 & 57 & 0.0025 & 55 & 0.0075 & 56 & 0.0066 \\
\textbf{Fathom flat 64} & 74 & 0.0021 & 75 & 0.0012 & 73 & 0.0043 & 74 & 0.0038 \\
\textbf{Fathom flat 80} & 92 & 0.0012 & 93 & 0.0007 & 91 & 0.0025 & 92 & 0.0026 \\
\textbf{Fathom flat 128} & 146 & 0.0003 & 148 & 0.0002 & 146 & 0.0010 & 146 & 0.0013 \\
SparQ $r{=}16$ & 68 & 0.0071 & 68 & 0.0044 & 68 & 0.0090 & 68 & 0.0091 \\
SparQ $r{=}32$ & 136 & 0.0022 & 136 & 0.0016 & 136 & 0.0025 & 136 & 0.0020 \\
Double Sparsity $c{=}32$ & 136 & 0.0030 & 136 & 0.0015 & 136 & 0.0043 & 136 & 0.0042 \\
Loki $r{=}32$ & 136 & 0.0155 & 136 & 0.0008 & 136 & 0.0026 & 136 & 0.0024 \\
Loki $r{=}64$ & 272 & 0.0004 & 272 & 0.0001 & 272 & 0.0002 & 272 & 0.0002 \\
2-bit thumbnail & 288 & 0.0356 & 288 & 0.0002 & 288 & 0.0025 & 288 & 0.0014 \\
full 4-bit scan & 544 & $<$0.0001 & 544 & $<$0.0001 & 544 & $<$0.0001 & 544 & $<$0.0001 \\
\textbf{Fathom KLT 40} & 47 & 0.0214 & 47 & 0.0013 & 47 & 0.0065 & 47 & 0.0052 \\
\textbf{Fathom KLT 48} & 56 & 0.0167 & 57 & 0.0009 & 56 & 0.0044 & 56 & 0.0038 \\
\textbf{Fathom KLT 64} & 74 & 0.0102 & 75 & 0.0004 & 74 & 0.0025 & 74 & 0.0019 \\
\textbf{Fathom KLT 80} & 93 & 0.0060 & 93 & 0.0003 & 92 & 0.0017 & 92 & 0.0010 \\
\bottomrule
\end{tabular}}
\end{table*}

\begin{table*}[tbp]\centering\small
\caption{Qwen2.5-7B-Instruct-1M, 128k context. Error at three selection ratios $k/n$ (0.1\%, 0.4\%, 1.6\%).}
\label{tab:kratio}
\resizebox{0.8\textwidth}{!}{\begin{tabular}{lrlrlrl}
\toprule
 & \multicolumn{2}{c}{Qwen2.5-7B-1M 128k/$K{=}128$} & \multicolumn{2}{c}{Qwen2.5-7B-1M 128k/$K{=}512$} & \multicolumn{2}{c}{Qwen2.5-7B-1M 128k/$K{=}2048$} \\
 & bits & error & bits & error & bits & error \\
\midrule
\textbf{Fathom KLT 48} & 56 & 0.0196 & 56 & 0.0048 & 56 & 0.0013 \\
\textbf{Fathom KLT 64} & 74 & 0.0111 & 74 & 0.0025 & 74 & 0.0006 \\
\textbf{Fathom KLT 80} & 92 & 0.0070 & 92 & 0.0016 & 92 & 0.0003 \\
\textbf{Fathom raw 48} & 55 & 0.0300 & 55 & 0.0082 & 55 & 0.0025 \\
\textbf{Fathom raw 64} & 73 & 0.0162 & 73 & 0.0036 & 73 & 0.0013 \\
SparQ $r{=}16$ & 68 & 0.0303 & 68 & 0.0116 & 68 & 0.0055 \\
SparQ $r{=}32$ & 136 & 0.0064 & 136 & 0.0022 & 136 & 0.0011 \\
Double Sparsity & 136 & 0.0153 & 136 & 0.0052 & 136 & 0.0024 \\
Loki $r{=}32$ & 136 & 0.0132 & 136 & 0.0033 & 136 & 0.0013 \\
Loki $r{=}64$ & 272 & 0.0015 & 272 & 0.0004 & 272 & 0.0001 \\
2-bit thumbnail & 288 & 0.0131 & 288 & 0.0022 & 288 & 0.0001 \\
full 4-bit scan & 544 & 0.0002 & 544 & $<$0.0001 & 544 & $<$0.0001 \\
\bottomrule
\end{tabular}}
\end{table*}

\begin{figure}[tbp]\centering
\fig{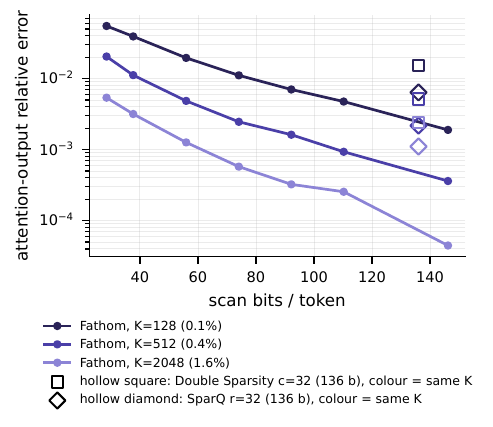}{0.85\linewidth}{selection ratio}
\caption{Qwen2.5-7B-Instruct-1M at 128k: error versus bits at three selection ratios. Hollow markers are Double Sparsity (square) and SparQ $r=32$ (diamond) at 136 bits in the colour of the same $k$.}
\label{fig:ratio}
\end{figure}

\begin{figure}[tbp]\centering
\fig{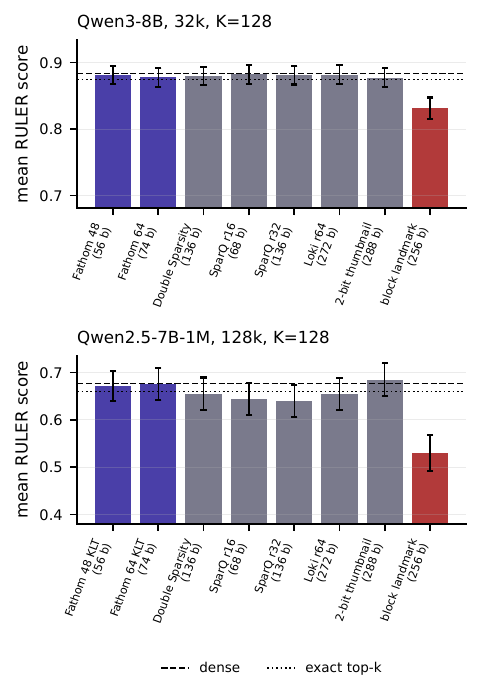}{\columnwidth}{RULER bars}
\caption{RULER-style mean score per method with bits in parentheses. Dashed: dense; dotted: exact top-$k$ oracle. (a) Qwen3-8B, 32k, $k=128$. (b) Qwen2.5-7B-Instruct-1M, 128k, $k=128$.}
\label{fig:ruler}
\end{figure}

\begin{figure}[tbp]\centering
\fig{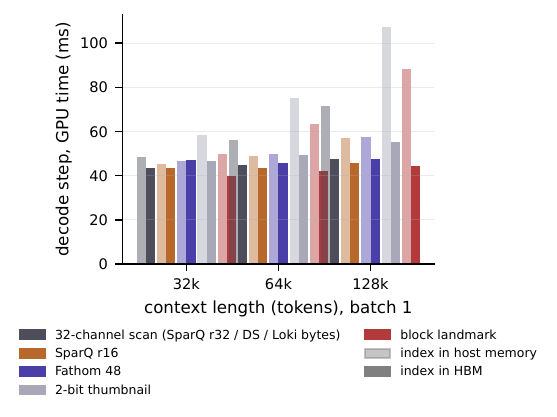}{0.85\linewidth}{index in HBM control}
\caption{Real prefill at 32k--128k: GPU time per step with the scan index in host memory versus in HBM. With every index resident in HBM ours is not faster than the channel scans or the landmark index.}
\label{fig:hbm}
\end{figure}

\begin{table}[tbp]\centering\small
\caption{Triton scan kernels on the A100 with the stores in HBM: Qwen3-8B, batch 4 $\times$ 8 KV heads, $k=512$, layers 0, 1, 2, 18 and 30, CUDA-graph replays, median of 30. Ours uses the per-layer plan, so its bits are the mean over these five layers, which include the early layers with the largest budgets. Scan time per layer and achieved HBM bandwidth.}
\label{tab:kernel}
\resizebox{\columnwidth}{!}{\begin{tabular}{lrrrrrr}
\toprule
 & \multicolumn{3}{c}{32k tokens} & \multicolumn{3}{c}{128k tokens} \\
method & bits & scan ms & GB/s & bits & scan ms & GB/s \\
\midrule
Fathom 48 & 84 & 0.140 & 75 & 84 & 0.415 & 102 \\
Fathom 64 & 95 & 0.155 & 77 & 95 & 0.457 & 106 \\
SparQ $r{=}16$ (16 channels) & 68 & 0.060 & 148 & 68 & 0.164 & 218 \\
32-channel scan & 136 & 0.102 & 175 & 136 & 0.291 & 245 \\
full 4-bit scan & 544 & 0.330 & 217 & 544 & 1.010 & 282 \\
\bottomrule
\end{tabular}}
\end{table}

\begin{figure}[tbp]\centering
\fig{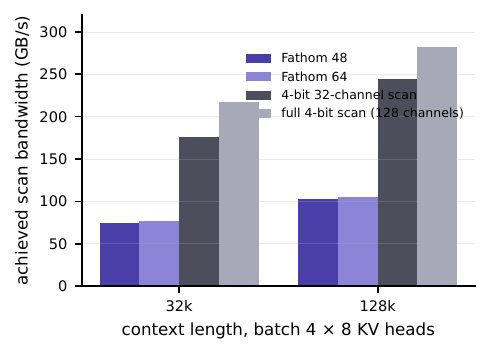}{0.85\linewidth}{A100 kernel bandwidth}
\caption{Achieved HBM bandwidth of the A100 scan kernels at 32k and 128k tokens, batch 4. The planes kernel reaches a third to a half of the channel scans' bandwidth; none of them approaches the 2\,TB/s peak.}
\label{fig:kernel}
\end{figure}

\begin{figure}[tbp]\centering
\fig{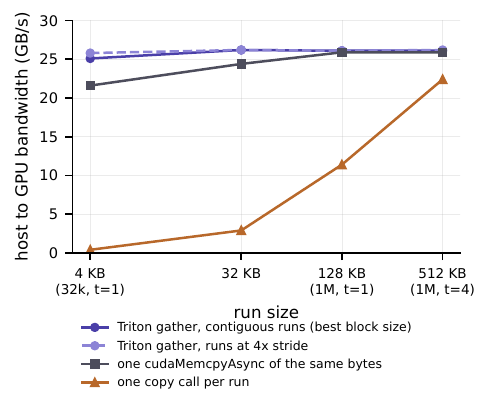}{0.85\linewidth}{PCIe access pattern}
\caption{PCIe transfer bandwidth versus run size on the A100 host: the Triton gather of contiguous runs, a single memcpy of the same bytes, and per-run copy calls.}
\label{fig:pcie}
\end{figure}

\begin{figure}[tbp]\centering
\fig{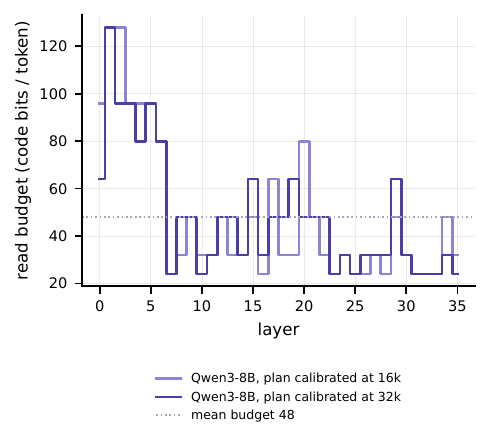}{0.85\linewidth}{per-layer budgets}
\caption{Per-layer bit budgets (code bits, before scales) for Qwen3-8B at mean 48, calibrated at 16k and at 32k.}
\label{fig:layers}
\end{figure}

\Needspace*{10\baselineskip}
\section{Additional ablations}\label{app:abl}
\paragraph{B1. Store precision.} With a per-query depth read the offline allocation of the store hardly matters (Table~\ref{tab:store}). We use the uniform 4-bit store, which is a 4-bit copy of K and needs no allocation step.
\begin{table}[tbp]\centering\small
\caption{B4: SparQ under its published grouped-query rule (one channel set per KV head) and a per-head variant that reads the union of the heads' picks.}
\label{tab:sparqvariant}
\resizebox{\columnwidth}{!}{\begin{tabular}{llrlrlrlrl}
\toprule
 & & \multicolumn{4}{c}{$r{=}16$} & \multicolumn{4}{c}{$r{=}32$} \\
 & & \multicolumn{2}{c}{published rule} & \multicolumn{2}{c}{per-head variant} & \multicolumn{2}{c}{published rule} & \multicolumn{2}{c}{per-head variant} \\
Model & ctx/$k$ & bits & error & bits & error & bits & error & bits & error \\
\midrule
Qwen3-8B & 16k/256 & 68 & 0.0077 & 156 & 0.0037 & 136 & 0.0022 & 298 & 0.0008 \\
Qwen3-8B & 32k/512 & 68 & 0.0080 & 156 & 0.0041 & 136 & 0.0012 & 296 & 0.0002 \\
Qwen3-8B & 32k/128 & 68 & 0.0305 & 156 & 0.0228 & 136 & 0.0073 & 296 & 0.0021 \\
Qwen3-4B & 16k/256 & 68 & 0.0071 & 156 & 0.0043 & 136 & 0.0022 & 299 & 0.0011 \\
Llama-3.1-8B & 4k/128 & 68 & 0.0044 & 148 & 0.0041 & 136 & 0.0016 & 284 & 0.0009 \\
Qwen2.5-7B & 32k/512 & 68 & 0.0090 & 200 & 0.0030 & 136 & 0.0025 & 373 & 0.0010 \\
Qwen2.5-7B & 32k/128 & 68 & 0.0238 & 200 & 0.0097 & 136 & 0.0068 & 373 & 0.0024 \\
Qwen2.5-7B-1M & 32k/512 & 68 & 0.0091 & 197 & 0.0024 & 136 & 0.0020 & 366 & 0.0002 \\
Qwen2.5-7B-1M & 128k/128 & 68 & 0.0303 & 192 & 0.0170 & 136 & 0.0064 & 362 & 0.0020 \\
Qwen2.5-7B-1M & 128k/512 & 68 & 0.0116 & 192 & 0.0054 & 136 & 0.0022 & 362 & 0.0008 \\
Qwen2.5-7B-1M & 128k/2048 & 68 & 0.0055 & 192 & 0.0020 & 136 & 0.0011 & 362 & 0.0003 \\
\bottomrule
\end{tabular}}
\end{table}
\begin{table}[tbp]\centering\small
\caption{B1: Qwen3-8B, 16k, $k=256$. The same read budgets over a uniform 4-bit store, a uniform 8-bit store and a rate-allocated 128-bit store.}
\label{tab:store}
\resizebox{\columnwidth}{!}{\begin{tabular}{lrlrlrl}
\toprule
 & \multicolumn{2}{c}{4-bit store} & \multicolumn{2}{c}{8-bit store} & \multicolumn{2}{c}{rate-allocated store} \\
read budget & bits & error & bits & error & bits & error \\
\midrule
mean 48 & 56 & 0.0052 & 56 & 0.0052 & 56 & 0.0054 \\
mean 64 & 74 & 0.0024 & 74 & 0.0024 & 74 & 0.0026 \\
mean 80 & 92 & 0.0012 & 92 & 0.0012 & 92 & 0.0017 \\
\bottomrule
\end{tabular}}
\end{table}

\paragraph{B2. Loki's rank.} Loki's failure on Qwen3-8B is rank, not quantization: fp16 coordinates at $r=32$ do not repair it and $r=64$ at 4 bits does (Table~\ref{tab:loki}). On Llama-3.1-8B at 4k $r=32$ is fine. The failure is model- and length-specific, which is why the equal-error references are Double Sparsity and SparQ $r=32$.
\begin{table}[tbp]\centering\small
\caption{B2: Loki at rank 32 in fp16 and 4-bit, and at rank 64.}
\label{tab:loki}
\resizebox{\columnwidth}{!}{\begin{tabular}{lllll}
\toprule
Model & ctx/$k$ & $r{=}32$ fp16 (512 b) & $r{=}32$ 4-bit (136 b) & $r{=}64$ 4-bit (272 b) \\
\midrule
Qwen3-8B & 16k/256 & 0.0105 & 0.0109 & 0.0004 \\
Qwen3-8B & 32k/512 & 0.4007 & 0.4033 & 0.0008 \\
Qwen3-8B & 32k/128 & 0.6749 & 0.6898 & 0.0110 \\
Llama-3.1-8B & 4k/128 & 0.0008 & 0.0008 & 0.0001 \\
\bottomrule
\end{tabular}}
\end{table}

\paragraph{B3. Active channels.} Table~\ref{tab:active} gives the measured number of channels a read touches at each budget.
\begin{table}[tbp]\centering\small
\caption{B3: mean number of active channels (of 128) per query read, flat budgets, headline settings.}
\label{tab:active}
\resizebox{\columnwidth}{!}{\begin{tabular}{llrrrrr}
\toprule
Model & ctx/$k$ & mean 32 & mean 48 & mean 64 & mean 80 & mean 128 \\
\midrule
Qwen3-8B & 16k/256 & 22 & 31 & 40 & 48 & 73 \\
Qwen3-8B & 32k/512 & 21 & 30 & 39 & 48 & 73 \\
Qwen3-4B & 16k/256 & 21 & 31 & 40 & 49 & 74 \\
Llama-3.1-8B & 4k/128 & 24 & 34 & 44 & 54 & 79 \\
Qwen2.5-7B & 32k/512 & 22 & 31 & 40 & 49 & 74 \\
Qwen2.5-7B-1M & 32k/512 & 23 & 32 & 40 & 49 & 74 \\
Qwen2.5-7B-1M & 128k/2048 & 23 & 32 & 40 & 49 & 72 \\
\bottomrule
\end{tabular}}
\end{table}

\paragraph{B5. Calibration domain.} Fathom's variances and KLT bases, and Double Sparsity's channel order, are calibrated on Wikitext-103; SparQ needs no calibration. Table~\ref{tab:calibfid} measures attention error on 128k tokens of held-out agent transcripts (rendered exactly as the sessions of \S\ref{sec:agent}, from repositories not used there) with the statistics taken from Wikitext and from a second set of agent transcripts. Matching the domain lowers Fathom's error by at most \calibWikiFortyEightFiveTwelve{} to \calibAgentFortyEightFiveTwelve{} at 56 bits and \calibWikiEightyFiveTwelve{} to \calibAgentEightyFiveTwelve{} at 92 bits, and Double Sparsity's by \calibDsWikiFiveTwelve{} to \calibDsAgentFiveTwelve; no rank against SparQ $r=32$ changes. Table~\ref{tab:agentcalib} repeats the $k=512$ agent decode with the agent-domain statistics and with a third option that removes offline calibration altogether: the mean, eigenbasis and eigenvalues of each layer's keys computed from the session's own prefill cache (one covariance per KV head, under a millisecond at 100k tokens). Both raise step agreement by \agentPairedAgentFortyEightWiki{} to \agentPairedAgentEightyWiki{} over the Wikitext calibration, within 1.5 standard errors, and the session-calibrated 92-bit read at \agentOsimplanesSeighty{} is \agentPairedSelfEightySparqThirtyTwo{} $\pm$ \agentPairedSESelfEightySparqThirtyTwo{} against SparQ $r=32$; recalibrating Double Sparsity changes it by \agentPairedDsAgentWiki. The exact top-$k$ step was reproduced verbatim in all \agentCalibOracleSame{} sessions between the two runs.
\begin{table}[tbp]\centering\small
\caption{B5: attention error on agent-transcript keys (Qwen2.5-7B-Instruct-1M, 128k) with statistics calibrated on Wikitext or on held-out agent transcripts. SparQ has no calibration.}
\label{tab:calibfid}
\resizebox{\columnwidth}{!}{\begin{tabular}{lrllll}
\toprule
 & & \multicolumn{2}{c}{$k{=}512$ (0.4\%)} & \multicolumn{2}{c}{$k{=}2048$ (1.6\%)} \\
method & bits & Wikitext calib. & agent calib. & Wikitext calib. & agent calib. \\
\midrule
\textbf{Fathom KLT 48} & 56 & 0.0057 & 0.0053 & 0.0018 & 0.0015 \\
\textbf{Fathom KLT 64} & 74 & 0.0030 & 0.0030 & 0.0008 & 0.0007 \\
\textbf{Fathom KLT 80} & 92 & 0.0017 & 0.0014 & 0.0003 & 0.0002 \\
SparQ $r{=}16$ & 68 & 0.0102 & 0.0102 & 0.0057 & 0.0057 \\
SparQ $r{=}32$ & 136 & 0.0025 & 0.0025 & 0.0011 & 0.0011 \\
Double Sparsity & 136 & 0.0050 & 0.0044 & 0.0023 & 0.0021 \\
\bottomrule
\end{tabular}}
\end{table}
\begin{table}[tbp]\centering\small
\caption{B5: the $k=512$ agent sessions of Table~\ref{tab:agent} with Fathom and Double Sparsity calibrated on Wikitext (as in the main text), on held-out agent transcripts, or on the session's own keys.}
\label{tab:agentcalib}
\resizebox{\columnwidth}{!}{\begin{tabular}{lrl}
\toprule
method & bits & step agreement with exact top-$k$, mean $\pm$ s.e. \\
\midrule
\textbf{Fathom 48} & 56 & \textbf{0.53} $\pm$ 0.04 \\
\textbf{Fathom 48, agent-calibrated} & 56 & \textbf{0.54} $\pm$ 0.04 \\
\textbf{Fathom 48, session-calibrated} & 56 & \textbf{0.57} $\pm$ 0.04 \\
\textbf{Fathom 64} & 74 & \textbf{0.54} $\pm$ 0.04 \\
\textbf{Fathom 64, agent-calibrated} & 74 & \textbf{0.60} $\pm$ 0.04 \\
\textbf{Fathom 64, session-calibrated} & 74 & \textbf{0.54} $\pm$ 0.05 \\
\textbf{Fathom 80} & 92 & \textbf{0.60} $\pm$ 0.05 \\
\textbf{Fathom 80, agent-calibrated} & 92 & \textbf{0.67} $\pm$ 0.04 \\
\textbf{Fathom 80, session-calibrated} & 92 & \textbf{0.64} $\pm$ 0.04 \\
Double Sparsity $c{=}32$ & 136 & 0.55 $\pm$ 0.04 \\
Double Sparsity $c{=}32$, agent-calibrated & 136 & 0.53 $\pm$ 0.04 \\
SparQ $r{=}32$ & 136 & 0.60 $\pm$ 0.04 \\
\bottomrule
\end{tabular}}
\end{table}

\Needspace*{14\baselineskip}
\section{Hardware and software}
A100-SXM4-80GB (RunPod secure cloud, PCIe 4.0 host link, 2\,TB host RAM, 16 vCPU), PyTorch 2.8, Triton 3.4, Transformers 4.56.1, CUDA 12.8. Llama-3.1-8B activations from an NVIDIA L4 (24\,GB). Kernels are Triton; the gather kernel copies contiguous runs from mapped pinned host memory into HBM staging. Every table and every number in the text is generated from the result files by one script.
\end{document}